\documentclass{article}
\PassOptionsToPackage{numbers}{natbib}
\usepackage[final]{neurips_2023}

\usepackage[utf8]{inputenc} 
\usepackage[T1]{fontenc}    
\usepackage{hyperref}       
\usepackage{url}            
\usepackage{booktabs}       
\usepackage{amsfonts}       
\usepackage{nicefrac}       
\usepackage{microtype}      
\usepackage{xcolor}         

\usepackage{subcaption}
\usepackage{graphicx}
\usepackage{color}
\usepackage{multirow}
\usepackage{wrapfig}

\usepackage{amssymb}
\usepackage{adjustbox}
\usepackage{tabularx}
\newcommand{\sst}[1]{{\color{black}#1}}

\title{Benchmarking Robustness of Text-Image Composed Retrieval 
}

\author{%
  Shitong Sun$^{1}$ \;
  Jindong Gu$^{2}$ \;
  Shaogang Gong$^{1}$ \;\\
  $^{1}$Queen Mary University of London \; 
  $^{2}$ University of Oxford\\
  \texttt{shitong.sun@qmul.ac.uk, jindong.gu@eng.ox.ac.uk, s.gong@qmul.ac.uk}\\
}

\begin{document}

\maketitle

\begin{abstract}
	Text-image composed retrieval aims to retrieve the target image through the composed query, which is specified in the form of an image plus some text that describes desired modifications to the input image.
	It has recently attracted attention due to its ability to leverage both information-rich images and concise language to precisely express the requirements for target images.
	%
	However, the robustness of these approaches against real-world corruptions or further text understanding has never been studied.
	In this paper, we perform the first robustness study and establish three new diversified benchmarks for systematic analysis of text-image composed retrieval against natural corruptions in both vision and text and further probe textural understanding.
	%
	%
	For natural corruption analysis, we introduce two new large-scale benchmark datasets, CIRR-C and FashionIQ-C for testing in open domain and fashion  domain respectively, both of which apply 15 visual corruptions and 7 textural corruptions.
	For textural understanding analysis, we introduce a new diagnostic dataset CIRR-D by expanding the original raw data with synthetic data, which contains modified text to better probe textual understanding ability including numerical variation, attribute variation, object removal, background variation, and fine-grained evaluation.
	The code and benchmark datasets are available at \url{https://github.com/SunTongtongtong/Benchmark-Robustness-Text-Image-Compose-Retrieval}.
	%
	%
\end{abstract}

\section{Introduction}
%

Text-image composed retrieval, known as composed image retrieval or text-guided image retrieval, attempts to retrieve an image of interest from gallery images through a composed query of a reference image and its corresponding modified text.
As a single word (`dog') can correspond to thousands of images (dogs in different breeds, poses, and scenarios), language is considered to be discrete and sparse, while images are regarded as dense and continuous.
Using both images and text as queries enables the effective utilization of the continuous and dense nature of images to accurately express the requirements while leveraging discrete and sparse text to bridge semantic gaps beyond what the images alone can capture.
It becomes fascinating for its potential in a wide range of real-world applications including fashion domain e-commerce~\cite{han2022fashionvil,han2023fame,goenka2022fashionvlp,han2022uigr,chen2020image} and open domain internet search~\cite{liu2021image,baldrati2022effective,gu2023compodiff,saito2023pic2word}.
However, existing text-image composed retrieval methods mostly are trained and tested on clean data, while the models in the real world may naturally encounter distribution shift ~\cite{wang2021measure}, such as text typos and image corruptions owing to weather change.
%
%
Additionally, there is currently no analysis of whether the model understands the meaning of the text  
rather than solely relying on finding correspondences with the main objects as a shortcut to the text-image composed retrieval task.
For example, with a source image of a dog and modified text `change to two dogs on the table', the model may retrieve the target image by only recognizing the word `dog' and `table' without the ability of numerical counting.
\sst{Whether text-image composed retrieval models are robust in real-world applications, where natural corruption exists in both images and text, remains unexplored.
Also, whether these models are robust across diverse textural understanding requirements, remains a domain blank.}


In this work, we make the first attempt to evaluate the robustness of text-image composed retrieval by building three new large-scale robustness benchmarks on both fashion and open domains.
%
We raise the following two questions: \textbf{Q1: }\emph{How robust are text-image composed retrieval models to natural corruption including both visual and textual?}
Further to evaluate the text understanding ability, we have the second question \textbf{Q2: } \emph{How robust are text-image composed retrieval models to text understanding?}

To answer the first question, we present two benchmark datasets on text-image composed retrieval task.
Based on two widely used datasets FashionIQ~\cite{guo2019fashion} in the fashion domain and CIRR~\cite{liu2021image} in the open domain, we propose our benchmark datasets, namely FashionIQ-C and CIRR-C, both with 15 visual corruptions and 7 textual corruptions to evaluate model robustness against natural corruption in both image and text.
To answer the second question, we introduce a new diagnostic dataset CIRR-D to probe text understanding abilities on five elementary scenarios including numerical variation, attribute manipulation, object removal, background variation, and fine-grained variation. 
%
In detail, we construct the diagnostic dataset by synthetic triplets based on the CIRR validation set and use existing triplets from the CIRR validation set with both main captions and extended captions.
Our experiments show that the new benchmarks we introduced are suitable for robustness analysis against natural corruption on both image and text and further probe text understanding ability.


Our contributions are:
(1) We make the first attempt to analyze the robustness of text-image composed retrieval methods against natural corruption (including visual and textual) and textual understanding (including five elementary variations).
(2) We introduce three new large-scale benchmarks including two benchmark datasets (FashionIQ-C and CIRR-C) to evaluate robustness against natural corruption in both image and text and one diagnostic benchmark CIRR-D to probe text understanding robustness.
(3) We present an empirical analysis and conduct extended experiments.

\vspace{-0.1cm}
\section{Related Works}
\vspace{-0.2cm}
Quantifying robustness aims to evaluate the model stability to defend against corruption including natural corruption~\cite{hendrycks2019robustness,chantry2022robustness,wang2021textflint}, adversarial attacks~\cite{croce2020robustbench,wang2021adversarial,wang2021textflint}, or to probe certain ability such as logical reasoning~\cite{sanyal2022robustlr} and visual content manipulation~\cite{li2020closer}.
%
%
Traditional works about robustness analysis mainly focus on single modality involving visual modality-based tasks 
like image classification~\cite{hendrycks2019robustness}, face detection~\cite{dooley2022robustness}; textual modality based task like text classification
~\cite{zeng2021certified} and audio modality based task like speech recognition~\cite{mitra2017robust}.
%
Recently, robustness analysis against multimodal tasks, which is closer to real life and attempts to take a step towards a reliable system, has appeared but is still in its infancy.
For example, Li et al.~\cite{li2020closer}  take the first step to systematically analyze the robustness of a multimodal task, Visual Question Answering (VQA), against 4 generic robustness including linguistic variation and visual content manipulation.
However, it is limited to VQA tasks and doesn't introduce benchmarks to pinpoint sophisticated reasoning abilities.
Schiappa et al.~\cite{chantry2022robustness} introduce natural corrupted visual and textual benchmarks on text-to-video retrieval.
However, the robustness analysis of the multimodal underlying hypothesis, which aims to generalize textual semantic and reasoning ability to visual space, is not discussed. 
We consider the analysis of both natural corruption in image and text and further underlying text understanding and take the first step to conduct an extensive analysis of the natural corruption and text understanding of the robustness of deep neural networks in text-image composed retrieval.
More related work about text-image composed retrieval and diagnostic analysis can be found in supplementary~\ref{supp:related_work}.

\begin{figure*}[t]
	\centering
	\includegraphics[width=0.95\textwidth]{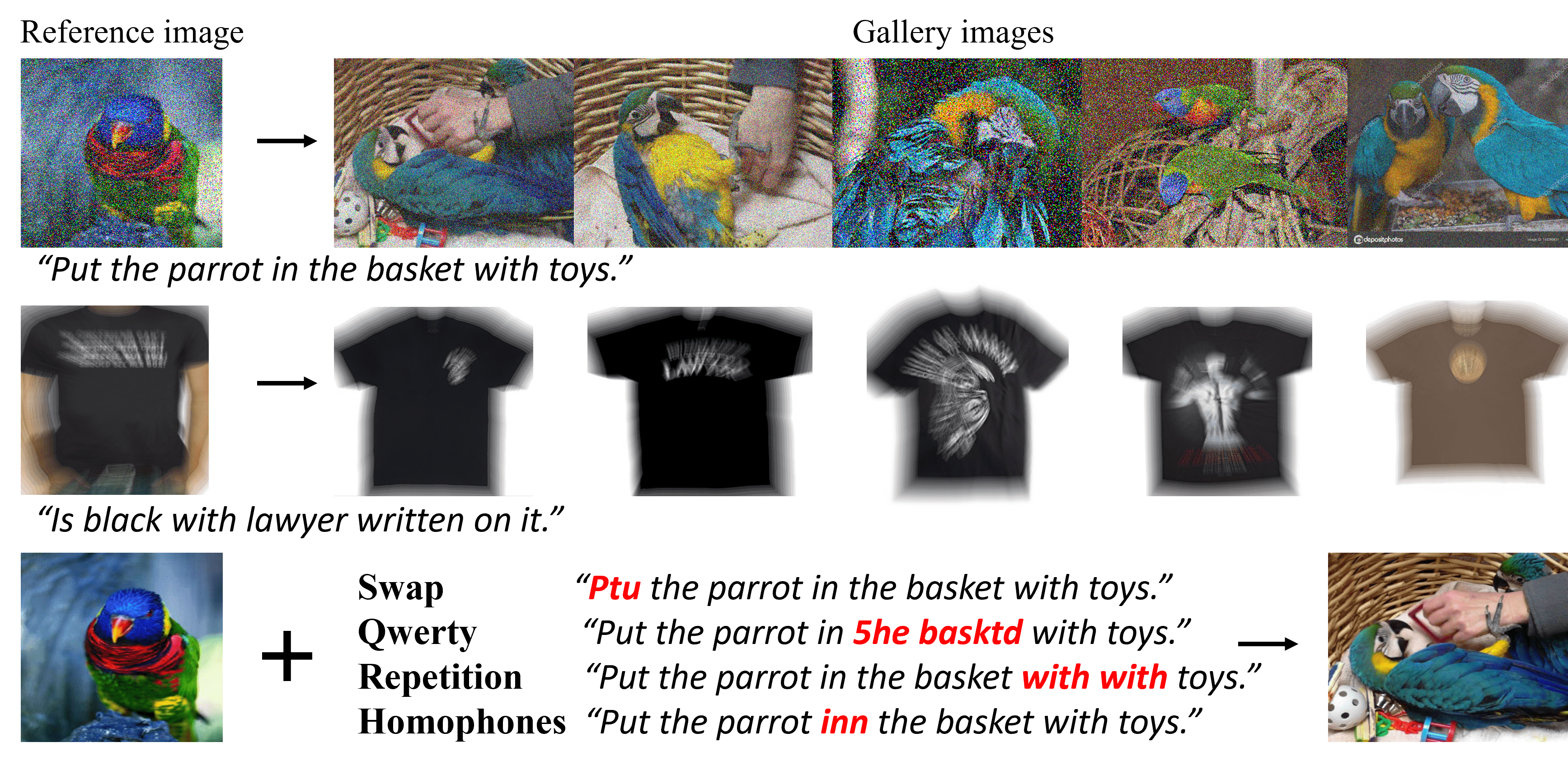}
	\caption{
		\sst
		{
			Sample visualization of proposed benchmarks under natural corruption with both visual and textual. \textbf{Top:} CIRR-C with impluse noise image corruption \textbf{Middle:} FashionIQ-C with zoom blur image corruption; \textbf{Bottom:} CIRR-C under character-level (Swap and Qwerty) and word-level (Repetition and Homophones) textual corruptions.
			%
			Gallery images are shown without particular order.
		}}
	\vspace{-0.3cm}
	\label{fig:samples_dataset}
\end{figure*}

\begin{figure*}[t]
	\centering
	\includegraphics[width=1.0\textwidth]{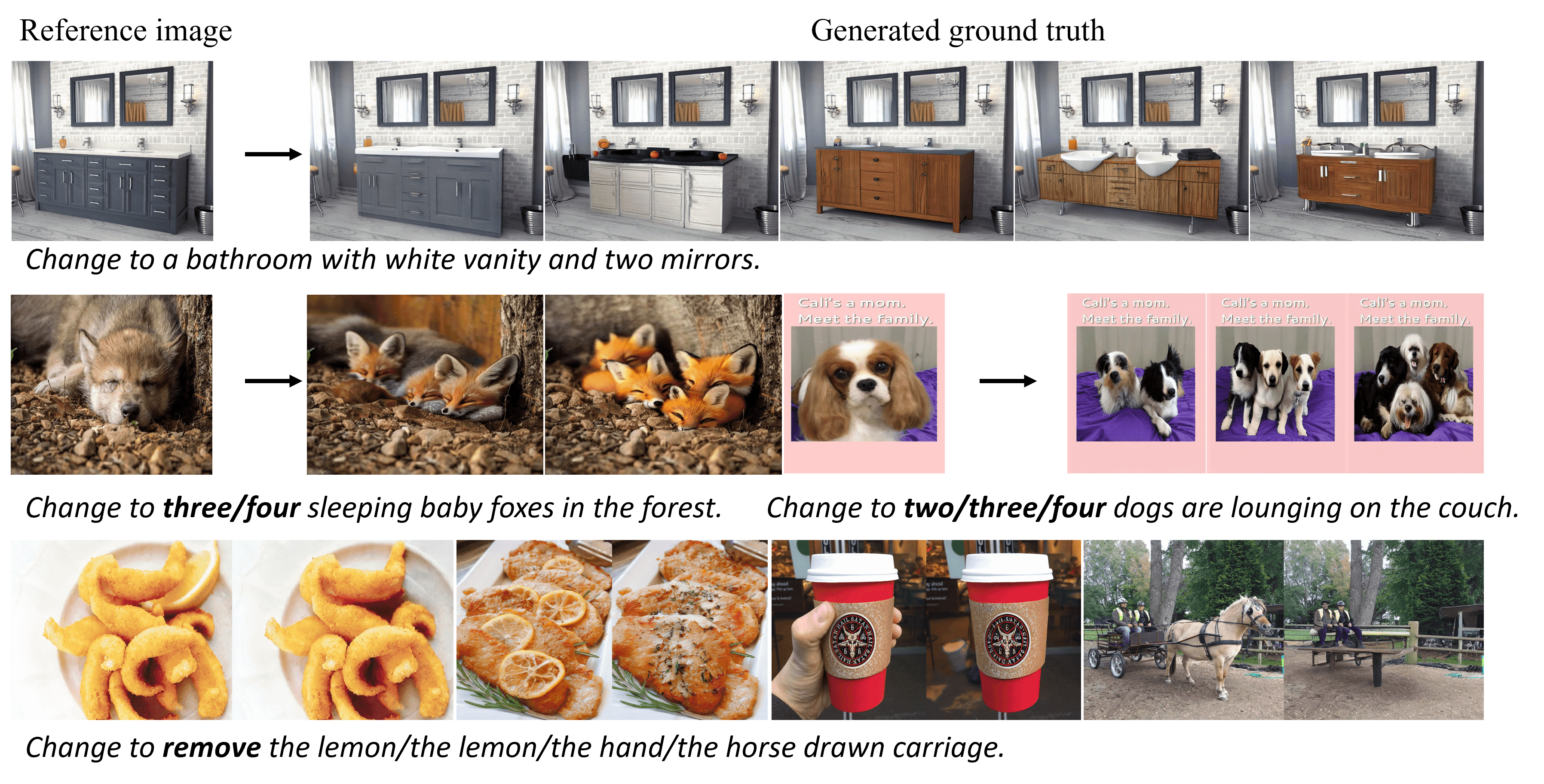}
	\caption{
		\sst
		{
			Sample visualization of proposed benchmarks probing text understanding. \textbf{Top:} CIRR-D with attribute variations; \textbf{Middle:} CIRR-D with number variations; \textbf{Bottom:} CIRR-D with object removal.
			%
			CIRR-D with background variation and fine-grained variation can be found in the supplementary materials.
		}}
	\vspace{-0.3cm}
	\label{fig:cirr_d_dataset}
\end{figure*}

\vspace{-5pt}
\section{Robustness Criteria for Text-Image Composed Retrieval}
\vspace{-5pt}

\textbf{Foundation of text-image composed retrieval.}
\label{sec:foundation}
Text-image composed retrieval aims to retrieve the target visual content through dense continuous images guided by sparse discrete text.
We discuss `dense' or `sparse' in the semantic space, where a single semantic word can correspond to thousands of images.
Therefore, text-image composed query can overcome the limitation of singular modality image retrieval, where text-image retrieval suffers from the unprecise description and unlimited correct targets, and image-image retrieval suffers from expression limitation without the ability to generalise to different visual content.
In light of this, the foundation abilities of text-image composed retrieval are threefold:
(1) Image representation to supply a precise anchor in the dense continuous visual space; 
(2) Text representation to supply subtle or significant differences between various visual contents, providing an unprecise target direction the model can generalize to; 
(3) Generalize sparse modified text attributes to dense reference images to precisely predict the target visual content by different variations of modality fusion. %
%

\textbf{Definition of robustness in text-image composed retrieval.}
%
According to the foundation of text-image composed retrieval above, a robust model should perform stable image feature extraction, text feature extraction, and modality fusion.
In light of this, the robustness of text-image composed retrieval can be defined in twofold: \emph{robustness against natural corruption} for both text and image and \emph{robustness against textual understanding} for consistent reasoning between textual and visual modality.
%
Specifically, for robustness against natural corruption, we evaluate text-image composed retrieval models under ubiquitous corruption frequently encountered in real-life in both visual and textual.
We evaluate 15 standard image corruptions with 5 severity categorized into noise, blur, weather, and digital following ~\cite{hendrycks2019robustness}.
We also evaluate 7 text corruptions categorized into character-level and word-level.
Further, for robustness against textual understanding, we evaluate common linguistic reasoning by selecting modified text with specific keywords or gallery set, categorized into
%
numerical variation, attributes variation, object removal, background variation, and fine-grained variation respectively.

\textbf{Evaluation metrics.}
To evaluate the performance of models in text-image composed retrieval, we adopt the standard evaluation metric in retrieval, i.e. Recall@K donated as R@K for short.
Further to measure robustness, we adopt relative robustness metrics $\gamma=1-\left(R_c-R_p\right) / R_c$ following ~\cite{chantry2022robustness,hendrycks2019robustness}, where $R_c$ and $R_p$ are the R@K under clean data and data with corruption respectively.
%
Additionally, in order to facilitate fair comparison among different models, we have expanded~\cite{delmas2022artemis} and established a unified testing platform for the convenient integration of various models.
In detail, we set gallery the whole validation set as in~\cite{delmas2022artemis, baldrati2022effective}, which includes more distractors and results to higher discriminative requirement, instead of setting gallery the same as the query set as in ~\cite{chen2020image,lee2021cosmo}.
Specifically for evaluating the fashionIQ dataset, we combine the two captions in a single query as ~\cite{baldrati2022effective,dodds2020modality} instead of combining the two modified captions in forward and reverse direction as ~\cite{lee2021cosmo}.
All the evaluated models are trained in three categories jointly and tested individually for dress, shirt, and toptee categories.
The reported results for fashionIQ are average of the three categories.

\textbf{Evaluation datasets.}
We utilize three new benchmarks for our text-image composed retrieval experiments, which are generated from two existing datasets: FashionIQ~\cite{guo2019fashion} in the fashion domain and CIRR~\cite{liu2021image} in the open domain.
Both datasets include human-generated captions that distinguish image pairs.
FashionIQ is based on the fashion domain containing 77,684 garment images, which can be divided into three categories: dress, shirt, and toptee.
Each image in FashionIQ contains a single subject positioned centrally with a clean background.
CIRR is composed of 21,552 real-life images extracted from NLVR2~\cite{suhr2018corpus}, which contains rich visual content in diverse backgrounds.
As shown in Figure.~\ref{fig:samples_dataset}, we build our benchmark and evaluate text-image composed retrieval models on text and image natural corruption robustness. 
Further as shown in Figure.~\ref{fig:cirr_d_dataset}, we expand the CIRR dataset and evaluate text understanding robustness. 

\emph{Natural vision and text corruption.}
\label{sec:natural}
To evaluate the robustness of the text-image composed retrieval model against natural corruption in both image and text, we create our robustness benchmark CIRR-C and FashionIQ-C with 15 visual corruptions and 7 textual corruptions.
For vision corruption, we follow~\cite{hendrycks2019robustness} to implement 15 standard natural corruptions which fall into four categories: noise, blur, weather, and digital, each having a severity from 1 to 5.
For text corruption, we follow~\cite{rychalska2019models} and implement the most related seven corruptions including four character-level corruptions and three word-level corruptions respectively.


\emph{Diagnostic datasets.}
\label{sec:diagnostic}
Following the current methods~\cite{liu2021image,baldrati2022effective} reporting the results on the validation set, we expand and build our probing datasets CIRR-D based on the validation set of CIRR to pinpoint text understanding ability including numerical variation, attributes composition, object removal, background transformation, and fine-grained evaluation.
We hypothesize that the model's corresponding capabilities can be evaluated when the modified text involves descriptions such as numbers, attributes, objects removal, or changing the background; and the ability to deal with fine-grained variations can be evaluated when the gallery images are highly similar following~\cite{liu2021image}.
In light of this, we build the triplets (reference image, modified text, and target image) of our probing dataset according to the appearances of specific keywords in modified text: "zero" to "ten", "number" for numerical query; color, shape and size for attribute query, "remove" for object removal query; "background" for background variation query.
As shown in Table~\ref{tab:synthetic}, the construction of CIRR-D dataset can fall into five probing categories from three sources as follows:
(1) The existing validation set of CIRR is composed of 2297 images and 4181 triplets, which is currently widely used.
Each image has a subset which is composed of 6 highly similar images as a gallery to better detect fine-grained discriminative ability.
(2) The auxiliary captions of the CIRR validation set, which is supplied but has not been used in the conventional evaluation.
These captions indicate the differences in removal content or background changes between image pairs, but they may not provide enough information to accurately locate the target image.
Therefore, we manually eliminated triplets that led to an excessive number of target images.
(3) Synthetic images we generate through Visual ChatGPT~\cite{wu2023visual} to bridge the language reasoning and visual recognition.
Based on the image from CIRR validation set, we generate its image caption by Visual ChatGPT and generate ten variants of the captions by ChatGPT including four for number variants, three for color variants, two for size variants, one for removing the objects respectively.
Afterward, based on the reference image and caption variants, Visual ChatGPT utilizes groundingDINO~\cite{liu2023grounding} to do object detection, segment anything~\cite{kirillov2023segany} to generate mask and stable diffusion~\cite{rombach2021highresolution} to generate target image.
We manually eliminated the unplausible generated images.
Our synthetic image pairs preserve the original background and only modify the specific areas mentioned in the text.

\textbf{Evaluated models.}
\label{sec:labels}
We perform our experiments on six text-image composed retrieval models. The modality fusion of these approaches is roughly summarized in Supplementary Table~\ref{tab:models_detail} including~\cite{vo2019composing,liu2021image,baldrati2022effective,delmas2022artemis,han2022fashionvil,dodds2020modality}, which can be divided into overlapped categories: (1) Large pretrained model: FashionViL, CIRPLANT, CLIP4CIR, whose pretrained dataset size are 1.35 million, 6.5 million, and 400 million image-text pairs respectively;
(2) Multi-task model: FashionViL, which is pretrained with four tasks at the same time;
(3) Light attention-based methods: ARTEMIS;
(4) Transformer-based models: MAAF, CIRPLANT, CIRPLANT and FashionViL.
(5) Lightweight models: MAAF, TIRG, and ARTEMIS (all with ResNet50 image encoder and LSTM text encoder for fair comparison).
%
(6) Single-modality models: Image-only (RN50) and Image-only(CLIP) are queried with images embedded by ResNet50 (same as evaluated TIRG, MAAF, ARTEMIS) and CLIP image encoder RN50x4 (same as evaluated CLIP4CIR) respectively. Text-only model is queried with text embedded using the CLIP text encoder. 
These methods were chosen because the reproduced results match with original reported results.
We test FashionViL~\cite{han2022fashionvil} in the fashion domain, CIRPLANT in the open domain, and all the rest published models in both the fashion domain and open domain.

\begin{table}[t!]
	\centering
	\scriptsize
	\caption{Relative robustness score for text-image composed retrieval under 15 natural image corruptions in CIRR-C Recall@10 and FashionIQ-C Recall@10. \sst{Recall@10 performance under clean conditions on the left.} \textbf{Bold} is the highest relative robustness for the five composed retrieval methods.}
	{\setlength\tabcolsep{2.2pt}
		\begin{tabular}{l|c|c c c|c c c c|c c c c|c c c c}
			\multicolumn{1}{c}{}        & \multicolumn{1}{c}{}                 & \multicolumn{3}{c}{Noise} & \multicolumn{4}{c}{Blur} & \multicolumn{4}{c}{Weather} & \multicolumn{4}{c}{Digital}                                                                                                                                                                                                                                                                          \\

			\toprule
			\textbf{CIRR-C}                   & \scriptsize{Clean}       & \scriptsize{Gauss.}       & \scriptsize{Shot}        & \scriptsize{Implu.}        & \scriptsize{Defoc.}        & \scriptsize{Glass}    & \scriptsize{Motion}   & \scriptsize{Zoom}      & \scriptsize{Snow}     & \scriptsize{Frost}    & \scriptsize{Fog}      & \scriptsize{Bright}   & \scriptsize{Contr.} & \scriptsize{Elast.}  & \scriptsize{Pixel}    & \scriptsize{JPEG}     \\

			\midrule
			Image-only(RN50)                   &  \sst{$50.4$}  & $0.57$                    & $0.55$                   & $0.58$                      & $0.68$                      & $0.28$                & $0.82$                & $0.45$                 & $0.38$                & $0.34$                & $0.64$                & $0.86$                & $0.20$                & $0.48$                & $0.76$                & $0.88$                \\
			Image-only(CLIP)                    &  \sst{$36.2$} & $0.56$                    & $0.55$                   & $0.58$                      & $0.66$                      & $0.32$                & $0.83$                & $0.49$                 & $0.52$                & $0.45$                & $0.77$                & $0.91$                & $0.24$                & $0.41$                & $0.78$                & $0.91$                \\

			Text-only(CLIP)                           & \sst{$51.2$}  & $0.79$                    & $0.76$                   & $0.81$                      & $0.85$                      & $0.29$                & $1.0$                 & $0.55$                 & $0.65$                & $0.70$                & $0.89$                & $1.0$                 & $0.19$                & $0.40$                & $0.96$                & $1.0$                 \\
			\midrule
			TIRG~\cite{vo2019composing}          & \sst{$55.1$} & $0.34$                    & $0.36$                   & $0.34$                      & $ 0.48$                     & $ 0.21$               & $ 0.70$               & $ 0.43$                & $ 0.31$               & $ 0.22$               & $0.40$                & $0.70$                & $0.12$                & $ 0.47$               & $0.74$                & $ 0.84$               \\
			MAAF~\cite{dodds2020modality}        &  \sst{$49.9$} & $0.50$                    & $0.49$                   & $0.50$                      & $0.62$                      & $0.26$                & $0.80$                & $0.41$                 & $0.36$                & $0.31$                & $0.50$                & $0.74$                & $0.11$                & $0.48$                & $0.83$                & $0.87$                \\
			ARTEMIS~\cite{delmas2022artemis}     & \sst{$59.0$} & $0.39$                    & $0.42$                   & $ 0.38$                     & $0.51$                      & $0.25$                & $ 0.70$               & $ 0.44$                & $ 0.31$               & $ 0.26$               & $0.45$                & $0.71$                & $0.10$                & $0.47$                & $0.75$                & $ 0.86$               \\

			CIRPLANT~\cite{liu2021image}         & \sst{$68.8$} & $\textbf{0.70}$           & $\textbf{0.69}$                   & $\textbf{0.71}$     & $0.77$                      & $0.28$                & $0.89$                & $0.51$                 & $0.44$                & $0.43$                & $0.66$                & $0.88$                & $\textbf{0.17}$                & $\textbf{0.56}$                & $0.85$                & $0.92$                \\

			CLIP4CIR~\cite{baldrati2022effective} & \sst{$\textbf{80.3}$} & $0.68$                    & $0.68$                   & $0.69$                      & $\textbf{0.77}$                      & $\textbf{0.28}$                & $\textbf{0.90}$                & $\textbf{0.52}$                 & $\textbf{0.55}$                & $\textbf{0.60}$                & $\textbf{0.80}$                & $\textbf{0.91}$                & $0.16$                & $0.39$                & $\textbf{0.91}$                & $\textbf{0.92}$                \\


			\bottomrule
		\end{tabular}}
		\label{tab:robustness_image}
		{\setlength\tabcolsep{2.4 pt}
		\begin{tabular}{l|c|c c c|c c c c|c c c c|c c c c}
			\multicolumn{1}{c}{}     & \multicolumn{1}{c}{}              & \multicolumn{3}{c}{Noise} & \multicolumn{4}{c}{Blur} & \multicolumn{4}{c}{Weather} & \multicolumn{4}{c}{Digital}                                                                                                                                                                                                                                          \\
			\toprule
			\textbf{FashionIQ-C}           & \scriptsize{Clean}         & \scriptsize{Gauss.}       & \scriptsize{Shot}        & \scriptsize{Implu.}        & \scriptsize{Defoc.}        & \scriptsize{Glass} & \scriptsize{Motion} & \scriptsize{Zoom} & \scriptsize{Snow} & \scriptsize{Frost} & \scriptsize{Fog} & \scriptsize{Bright} & \scriptsize{Contr.} & \scriptsize{Elast.} & \scriptsize{Pixel} & \scriptsize{JPEG} \\

			\midrule

			TIRG~\cite{vo2019composing}         & \sst{23.8} & $0.28$                    & $0.26$                   & $0.23$                      & $0.34$                      & $ 0.22$            & $  0.61$            & $0.57$            & $0.32$            & $0.27$             & $0.37$           & $0.61$              & $ 0.12$               & $ 0.64$              & $ 0.85$            & $ 0.85$           \\
			MAAF~\cite{dodds2020modality}       & \sst{23.4} & $0.31$                    & $0.27$                   & $0.25$                      & $0.44$                      & $0.21$             & $0.67$              & $0.53$            & $0.29$            & $0.24$             & $0.31$           & $0.54$              & $0.13$                & $0.54$               & $0.83$             & $0.83$            \\
			ARTEMIS~\cite{delmas2022artemis}    & \sst{24.9} & $ 0.24$                   & $0.24$                   & $0.20$                      & $ 0.38$                     & $0.26$             & $ 0.65$             & $ 0.60$           & $0.36$            & $ 0.25$            & $ 0.38$          & $ 0.55$             & $ 0.14$               & $ 0.63$              & $ 0.86$            & $ 0.87$           \\

			FashionViL~\cite{han2022fashionvil} & \sst{23.4} & $0.26$                    & $0.28$                   & $0.25$                      & $0.40$                      & $\textbf{0.31}$    & $\textbf{0.82}$     & $\textbf{0.67}$   & $0.33$            & $0.31$             & $0.34$           & $\textbf{0.70}$     & $0.15$                & $\textbf{0.86}$      & $\textbf{1.09}$    & $\textbf{1.06}$   \\
			CLIP4CIR~\cite{baldrati2022effective} &  \sst{\textbf{35.9}} & $\textbf{0.44}$           & $\textbf{0.42}$          & $\textbf{0.44}$             & $\textbf{0.54}$             & $0.21$             & $0.72$              & $0.50$            & $\textbf{0.46}$   & $\textbf{0.43}$    & $\textbf{0.60}$  & $\textbf{0.70}$     & $\textbf{0.22}$       & $0.37$               & $0.74$             & $0.83$            \\
			\bottomrule
		\end{tabular} }
\end{table}

\vspace{-0.3cm}

\section{Results and Analysis}
\vspace{-0.3cm}

\subsection{Natural corruption analysis}
\begin{figure}[t]
	\centering
	\includegraphics[width=1.0\textwidth]{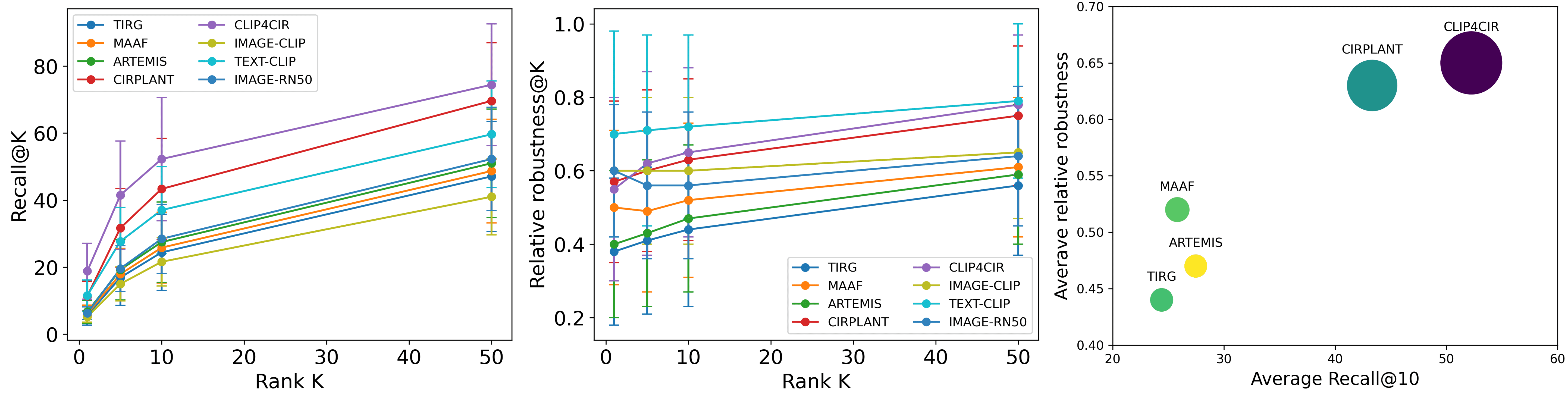}
	\caption{Models average performance in CIRR under 15 vision corruptions. Left: Recall vs. rank K. Middle: $\gamma$ vs. rank K. Right: $\gamma$ vs. recall@10, circle size indicates number of model parameters.} \vspace{-0.5cm}
	\label{fig:accuracies_compare}
\end{figure}

\vspace{-6pt}


In order to evaluate whether the text-image composed retrieval models are robust under natural corruptions, we perform experiments under 15 visual corruptions which can be further categorized into \textbf{noise}, \textbf{blur}, \textbf{weather}, \textbf{digital} corruptions on both fashion domain and open domain.
We report the relative robustness $\gamma$ under the highest severity of each natural visual corruption in Table~\ref{tab:robustness_image}, which shares the same trend as other corruption severities.
%
To evaluate the robustness against textual corruption, we conduct experiments under 7 textual corruptions on text-image composed retrieval models in \textbf{character-level} and \textbf{word-level}. Analysis about textual corruption is shown in supplementary~\ref{supp:textual}.
%

\textbf{Pretraining.}
Among the compared models, FashionViL CIRPLANT and CLIP4CIR are pretrained on large datasets, with respective sizes of 1.35 million, 6.5 million, and 400 million image-text pairs, while another three compared models are based on ImageNet pretrained ResNet50 as image encoder and random initialized LSTM as text encoder.
%
%
As shown in Figure~\ref{fig:accuracies_compare}, the models with large pretrained datasets consistently show better robustness 
in both open domain and fashion domain.
%
%
%
\emph{This implies that models with large pretrained datasets may result in better robustness against visual corruptions, which is in alignment with the statement from Paul et al.~\cite{paul2022vision}.}

\textbf{Bottleneck of robustness.}
We visualize the recall performance with rank K improvement on the left of Figure~\ref{fig:accuracies_compare} and relative robustness on the right.
With ranking $K$ improved, both the recall and relative robustness improved for all of the models.
Additionally, we find out in Figure~\ref{fig:accuracies_compare} that ResNet50-based image-only search shows even better accuracy as well as robustness compared with TIRG, ARTEMIS, and MAAF, which utilizes the same image encoder ResNet50.
According to the three foundations of text-image composed retrieval in Sec.~\ref{sec:foundation}, both the image encoder and modality fusion module can be vulnerable to vision corruption.
We observe that the ResNet50 backbone shows relatively high robustness while the modality fusion of the three models (TIRG, MAAF, and ARTEMIS) exacerbates the instability of the model.
However, this phenomenon is not applicable to the CLIP feature, whose text and image embedding are aligned in a unified space in the pretraining process.
Comparing Image-only (CLIP) and CLIP4CIR, which query with CLIP image embedding and CLIP text-image composed embedding respectively, we can find out CLIP4CIR consistently performs better recall performance as well as robustness in Figure~\ref{fig:accuracies_compare}.
\emph{Thus, we speculate that text features from aligned space can help boost the robustness, while text features from independent space will damage the model robustness.}

\subsection{Robustness against text understanding}
In this section, we analyze model reasoning ability through variation of modified text on numerical variation, attributes variation, object
removal, background variation, and fine-grained variation, which are supplied by our proposed CIRR-D dataset.
We take the 4181 queries from the origin CIRR dataset and evaluate them in our expanded CIRR-D dataset as a baseline, which incorporates diverse reasoning instruction and can represent the models' average performance across various instructions.
However, unlike other function detection, the gallery set for fine-grained is a subset, which is composed of six highly similar images following ~\cite{liu2021image}.
We first evaluate performance on each query type as shown in Table~\ref{tab:reasoning}.
Detailed analyses are discussed below.



\textbf{Numerical variation.}
To probe the ability of numerical variation, the modified text contains either a precise value of the number from zero to ten or an estimated value by comparison like `\emph{reduce/increase the number}'.
Comparing numerical specific query with CIRR query as shown in Table~\ref{tab:reasoning}, we can not observe significant variation which may result from the long-tailed distribution.
Namely, the numerical set has a large number of samples in the range of 1 to 3, while a small number of samples in the range of 4 to 10.
More analysis can be found in the supplementary~\ref{sec:limitation}.
%
%
For now, we speculate that \emph{numerical modification may not be the bottleneck of the current text-image composed retrieval.}  

%

\begin{table}[t!]
	\centering
	\small
	\caption{Recall of CIRR-D dataset. The red and green arrows indicate the performance increase or decrease compared with CIRR queries. \textbf{Bold} and \underline{underline} are
		the largest decrease and increase.  }
	{\setlength\tabcolsep{3pt}
		\begin{tabular}{l|c c c c c|c}
			\multicolumn{1}{c}{}                  & \multicolumn{5}{c}{R@5} & \multicolumn{1}{c}{Rsub@1}                                                                                                                                                                                                                          \\

			\toprule
			                                      & CIRR                    & Numerical                                                & Attribute                                               & Removal                                                 & Background                                            & Fine grained \\
			\midrule
			Image-only(RN50)                  & 31.55                   & 31.47 \textcolor{green}{$\downarrow$} (0.08)             & 32.57 \textcolor{red}{$\uparrow$} (1.02)                & 35.99 \textcolor{red}{$\uparrow$}   (4.44)              & 39.15 \textcolor{red}{$\uparrow$}  (\underline{7.60}) & 20.25        \\

			Image-only(CLIP)                      & 22.51                   & 24.80  \textcolor{red}{$\uparrow$} (2.29)                & 29.09 \textcolor{red}{$\uparrow$} (6.58)                & 27.90 \textcolor{red}{$\uparrow$} (\underline{5.39})    & 25.64 \textcolor{red}{$\uparrow$} (3.13)              & 20.02        \\
			Text-only                             & 39.02                   & 42.84  \textcolor{red}{$\uparrow$} (\underline{3.82})    & 49.45 \textcolor{red}{$\uparrow$} (\underline{10.43})   & 11.62 \textcolor{green}{$\downarrow$} (27.4)            & 11.62 \textcolor{green}{$\downarrow$} (\textbf{27.4}) & 53.73        \\
			\midrule
			TIRG~\cite{vo2019composing}           & 36.35                   & 39.64 \textcolor{red}{$\uparrow$} (3.29)                 & 37.77  \textcolor{red}{$\uparrow$} (1.42)               & 30.41  \textcolor{green}{$\downarrow$} (5.94)           & 32.82  \textcolor{green}{$\downarrow$} (3.53)         & 35.90        \\

			MAAF~\cite{dodds2020modality}         & 32.19                   & 32.53   \textcolor{red}{$\uparrow$} (0.34)               & 35.57 \textcolor{red}{$\uparrow$} (3.38)                & 31.09  \textcolor{green}{$\downarrow$} (1.10)           & 34.27  \textcolor{red}{$\uparrow$} (2.08)             & 28.63        \\
			ARTEMIS~\cite{vo2019composing}        & 40.05                   & 39.56   \textcolor{green}{$\downarrow$} (0.49)           & 42.68   \textcolor{red}{$\uparrow$} (2.63)              & 33.26  \textcolor{green}{$\downarrow$} (6.79)           & 35.56   \textcolor{green}{$\downarrow$} (4.49)        & 40.80        \\
			CIRPLANT~\cite{liu2021image}          & 48.82                   & 45.07    \textcolor{green}{$\downarrow$} (\textbf{3.75}) & 47.73   \textcolor{green}{$\downarrow$} (\textbf{1.09}) & 41.12  \textcolor{green}{$\downarrow$} (7.70)           & 45.98    \textcolor{green}{$\downarrow$} (2.84)       & 38.19        \\
			CLIP4CIR~\cite{baldrati2022effective} & 62.94                   & 64.18  \textcolor{red}{$\uparrow$} (1.24)                & 69.15   \textcolor{red}{$\uparrow$} (6.21)              & 31.66  \textcolor{green}{$\downarrow$} (\textbf{31.28}) & 41.88  \textcolor{green}{$\downarrow$} (21.06)        & 62.66        \\

			\bottomrule
		\end{tabular} }
	\label{tab:reasoning}
\end{table}

\textbf{Attributes variation.}
To evaluate the model's discriminative ability when querying elementary attributes, the modified text includes variations of color, shape and size.
As observed in Table~\ref{tab:reasoning}, all of the methods (except CIRPLANT) achieve higher performance with attribute queries than with CIRR queries.
Additionally, the performance of CLIP based image-only model and CLIP4CIR have an obvious increment of over 6\% compared with their performance with CIRR queries, which have a strong ability of attribute recognition including color, shape, and size.
%
%
This implies that \emph{attribute is the one of main focuses during training and models gain strong attribute discriminative ability.}


\textbf{Object removal.}
Object removal is a convenient approach to describe the differences between images but is universally overlooked by current methods in text-image composed retrieval.
To probe the ability of object removal through CIRR-D, the modified text of the query explicitly contains the word \emph{'remove'}.
As shown in Table~\ref{tab:reasoning}, all of the five compared methods achieve their lowest performance in object removal with an average decrement of 10.6\% compared with the CIRR query.
In particular, CLIP4CIR has a drop of over 30\%, which may be a result of its static pretraining process by aligning only image text pairs without comparison between images.
Surprisingly, image-only methods can have an increment over CIRR queries, which illustrates that visual similarity can boost the robustness over object removal but the text condition over guidance the model decision.
This aligns with the foundation of the task: images are dense and continuous while text is sparse and discrete.
\emph{In the case of object removal, text guidance expands the possibility of the targets, which distracts the model and results in lower performance.}

\textbf{Background variation.}
To probe the robustness against background modifications, the modified text of the query explicitly includes the word "background".
We observe a similar phenomenon as in object removal, where the performance of compared models (except MAAF) decreases but the performance of image-only models increases compared to CIRR queries.
As the CIRR-D sample visualization in Figure~\ref{fig:cirr-D},  
we can observe that the background modification method is limited such as changing the background color or making the background blur, which can lead to unrelated targets by relying solely on the text itself.
We further speculate that \emph{a modified text leading to more satisfactory candidates may result in impaired outcomes.}

%


\textbf{Fine-grained variation.}
To probe the fine-grained variation discriminative ability, we utilize the subset in the CIRR dataset, where each image is retrieved from its subset composed of another five highly similar images.
As the gallery is different from the above reasoning function, the recall cannot be compared with CIRR query performance directly.
We can observe from Table~\ref{tab:reasoning} that image-only models perform similarly to random guessing and text-only by CLIP embedding can achieve an acceptable result.
Among the five compared methods, TIRG achieve the lowest performance which hypothesizes the slight adjustment in visual space is sufficient rather than exploring text deeply and establishing a fusion space.
This phenomenon indicates that it is difficult to distinguish between two states in continuous visual space.
In contrast, text can precisely define subtle differences due to its discrete nature.
%
%
We also speculate that \emph{a modified text offers accurate information while minimizing the number of feasible targets can enhance the model's discriminative ability. }
%


\vspace{-0.3cm}

\section{Conclusion}
\vspace{-0.2cm}
In this work, we proposed three robustness benchmarks for text-image composed retrieval including two for natural corruption in both image and text and one for probing textual understanding.
%
%
%
Concretely, we first introduced two benchmark datasets, CIRR-C and FashionIQ-C with natural corruption (both image and text) in the open domain and fashion domain respectively.
Further, we create benchmark CIRR-D to assess the text understanding including number, attribute, object removal, background, and fine-grained variation.
\sst{Based on our observation, we provide the following suggestions to enhance model robustness in text-image composed retrieval:} 1) model pretrained on large datasets with little distribution shift will lead to better robustness, 2) text features from aligned space can help boost the robustness, while text features from independent space will damage the model robustness,
3) a modified text is more likely to enhance the model's discriminative ability when it minimizes the number of feasible targets and will distract the model when it leads to more satisfactory candidates.
These findings can potentially boost the robustness of text-image composed retrieval in the future.
\bibliographystyle{plain}

\bibliography{mybib}
\newpage

\appendix

\section{Creating Benchmark Datasets}
\label{sec:project_url}
We build three benchmark datasets for this work evaluating both natural corruption (CIRR-C and FashionIQ-C) and textual understanding (CIRR-D).
To evaluate natural corruption with both image and text, we introduce CIRR-C and FashionIQ-C based on the existing dataset CIRR and FashionIQ.
To evaluate textual understanding including variations of numerical, attributes (colour, shape and size), object removal, background and fine-grained details, we introduce CIRR-D by categorizing and expanding CIRR with synthetic images.
We provide raw images and complete code for generating all types of natural corruptions and the evaluation testbed in our code zip file.
For shortcut, we provide raw image link for \href{https://drive.google.com/drive/folders/1_D8vaSlLHl-5FUJxI8_mnEFWWzu-4_Dz?usp=drive_link}{CIRR}, \href{https://drive.google.com/drive/folders/14JG_w0V58iex62bVUHSBDYGBUECbDdx9}{FashionIQ} and \href{https://drive.google.com/drive/folders/1L119pdiG7aC_Z48oh_x4h3oXmz0faxV9?usp=sharing}{CIRR-D}.
To implement \textbf{CIRR-D} dataset, both raw images and queries in different categories (numerical, attribute, object removal, background and fine-grained variations) are provided directly.
To implement \textbf{CIRR-C} and \textbf{FashionIQ-C} dataset, research can recreate the same benchmark datasets with the following steps:

1. Download CIRR and FashionIQ raw images with our provided link.

2. Preprocess image or text with the provided code of image corruption and text corruption.

3. Apply the proposed corruptions with our testbed for downstream model evaluation.

\textbf{Evaluation settings.}
In order to ensure the fairness of the evaluation, we establish a standardized testbed for variant models except FashionViL to unify the evaluation process.
Further to reproduce the original performance, we implement the official pre-trained weight for FashionViL and CLIP4CIR.
We retrain the model and receive similar results as reported for models MAAF, TIRG, ARTEMIS, and CIRPLANT.
\sst{We implement zero-shot evaluation for BLIP2 and instructBLIP. And train BLIP2-CIR following~\cite{baldrati2022effective}.}
In detail, we extend the existing ARTEMIS code framework to facilitate the convenient interface of different trained models, where TIRG and ARTEMIS are already implemented.
For image input among these models, CIRPLANT is based on frozen ResNet152 pre-trained features while other models take raw images as input.
We implement a frozen ResNet152 image encoder so that we can introduce corruption to the raw image directly.
\sst{We leverage OPT-2.7b captioner for BLIP2 and vicuna-7b for instructBLIP.}
For all the 15 image corruptions, we perform the highest severity of corruption for obvious performance.
For text input among these models, TIRG, MAAF, and ARTEMIS build the vocabulary based on appearance words in target evaluation caption, while FashionViL, CIRPLANT, and CLIP4CIR implement their vocabulary from large pretraining dataset.
We implement textual corruption and change the raw text directly.
For a concise explanation, we show the result of FashionIQ by showing the average of the three categories: dress, shirt, and toptee.

\textbf{Implementation details}
\label{sec:implementation}
We supply 5 models (TIRG, MAAF, ARTEMIS, CIRPLANT and CLIP4CIR) in the same testbed to have a fair comparison with different benchmark datasets.
The selection of models or datasets can be easily accomplished through input parameters.
Further models can be implemented in our testbed by simply providing model structure files with the necessary interface.
In detail, the necessary interfaces include image feature extraction, text feature extraction, feature composing process and distance comparison.
Our testbed is compatible with two environments and five models.
CIRPLANT is implemented with Python(3.1) and Pytorch(1.8.1).
TIRG, MAAF, CLIP4CIR and ARTEMIS were implemented with Python(3.8) and Pytorch(2.0).
All the experiments are conducted and trained on NVIDIA A100 GPUs.
We will maintain our code for benchmarking and testbed future in GitHub.
Any questions about benchmark creation or evaluation can be raised by an issue on the GitHub page.

\section{Sample visualization}
\label{sec:visualization}

\subsection{CIRR-C visualization}
\begin{figure}[htbp]
	\centering
	\begin{minipage}[t]{1.0\textwidth}
		\centering
		\includegraphics[width=\textwidth]{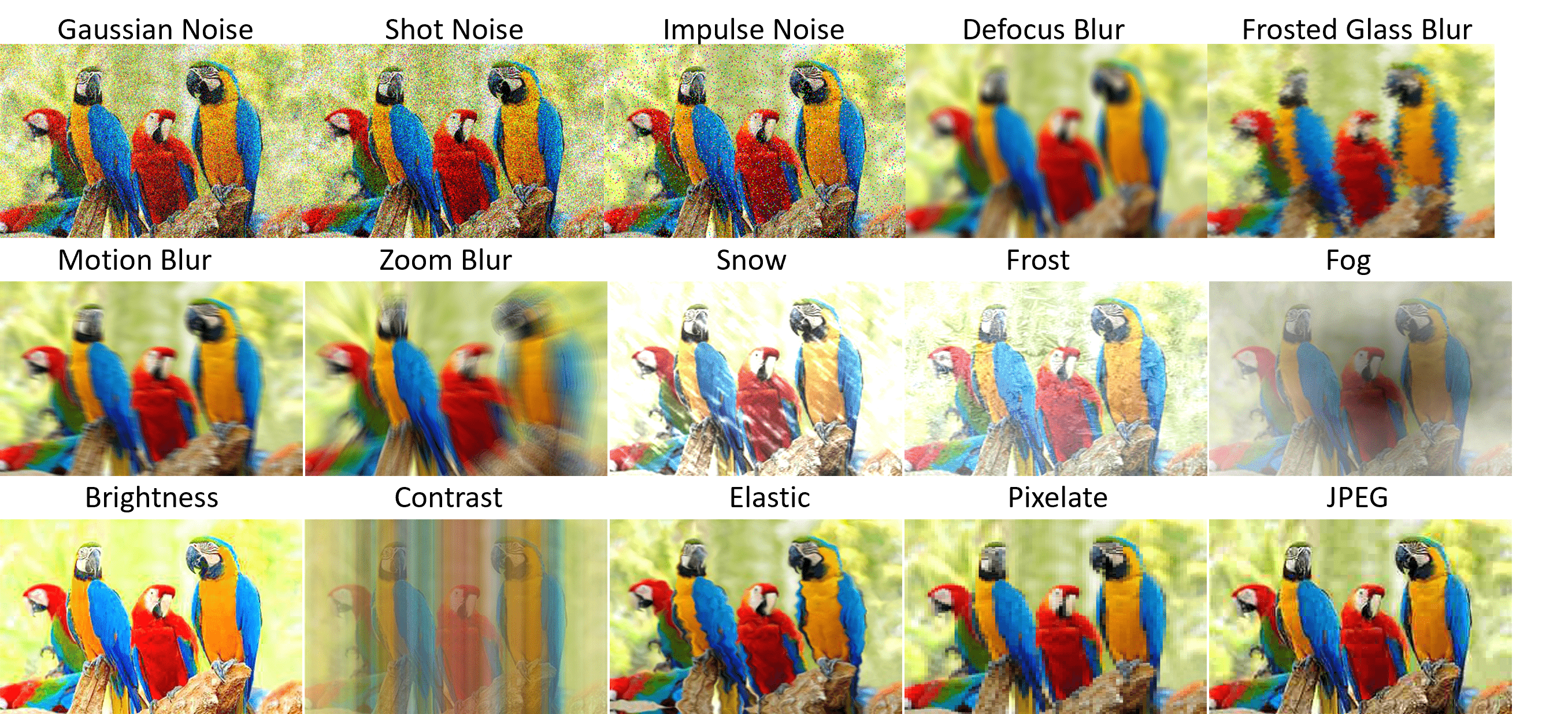}
		(a) Sample visualization with 15 standard image corruptions.
	\end{minipage}
	\begin{minipage}[t]{1.0\textwidth}
		\centering
		\includegraphics[width=\textwidth]{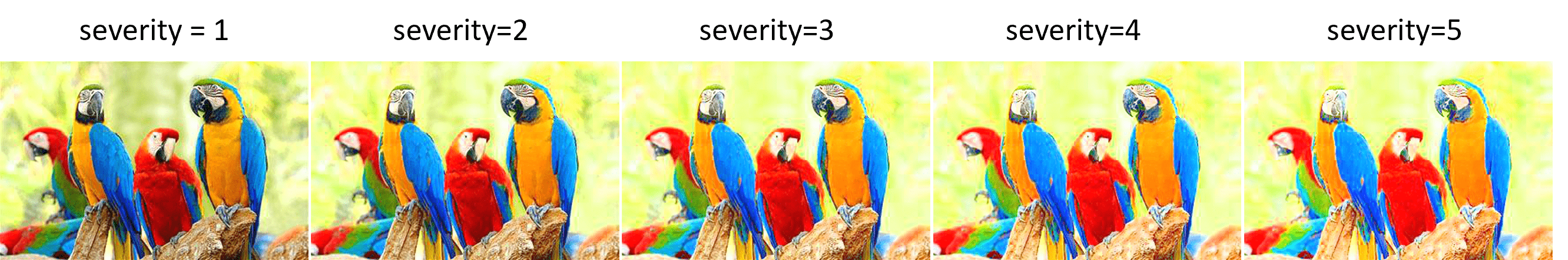}
		(b) Sample visualization of brightness corruption with 5 severities.
	\end{minipage}
	\begin{minipage}[t]{1.0\textwidth}
		\centering
		\includegraphics[width=\textwidth]{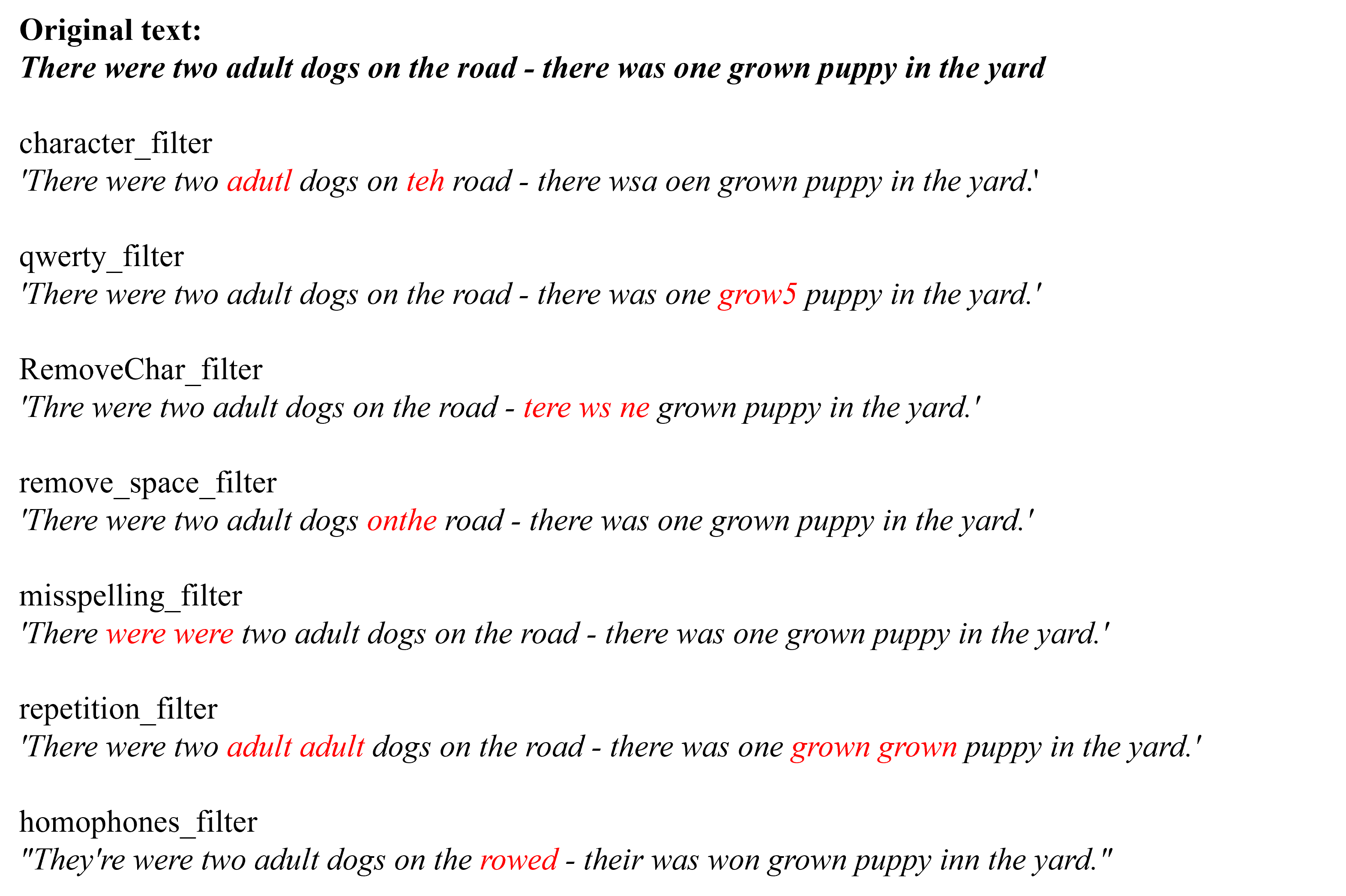}
		(c) Sample visualization of 7 text corruptions.
	\end{minipage}

	\caption{CIRR-C sample visualization: (a) 15 standard image corruptions, (b) 5 severities of brightness corruption and (c) 7 text corruption.}
	\label{fig:CIRR-C}
\end{figure}

We show the visualization samples from CIRR-C in Figure.~\ref{fig:CIRR-C}.
Our CIRR-C is based on the CIRR dataset and implemented with both image corruptions and text corruptions.
We apply 15 standard natural image corruptions, as depicted in Figure~\ref{fig:CIRR-C} (a), and demonstrate 5 levels of severity using brightness corruption as an example in Figure~\ref{fig:CIRR-C} (b).
We further visualize 7 text corruptions in Figure~\ref{fig:CIRR-C} (c).
For both image and text corruption, humans can easily recognize them.

\subsection{FashionIQ-C visualization}
\begin{figure}[htbp]
	\centering
	\begin{minipage}[t]{0.85\textwidth}
		\centering
		\includegraphics[width=\textwidth]{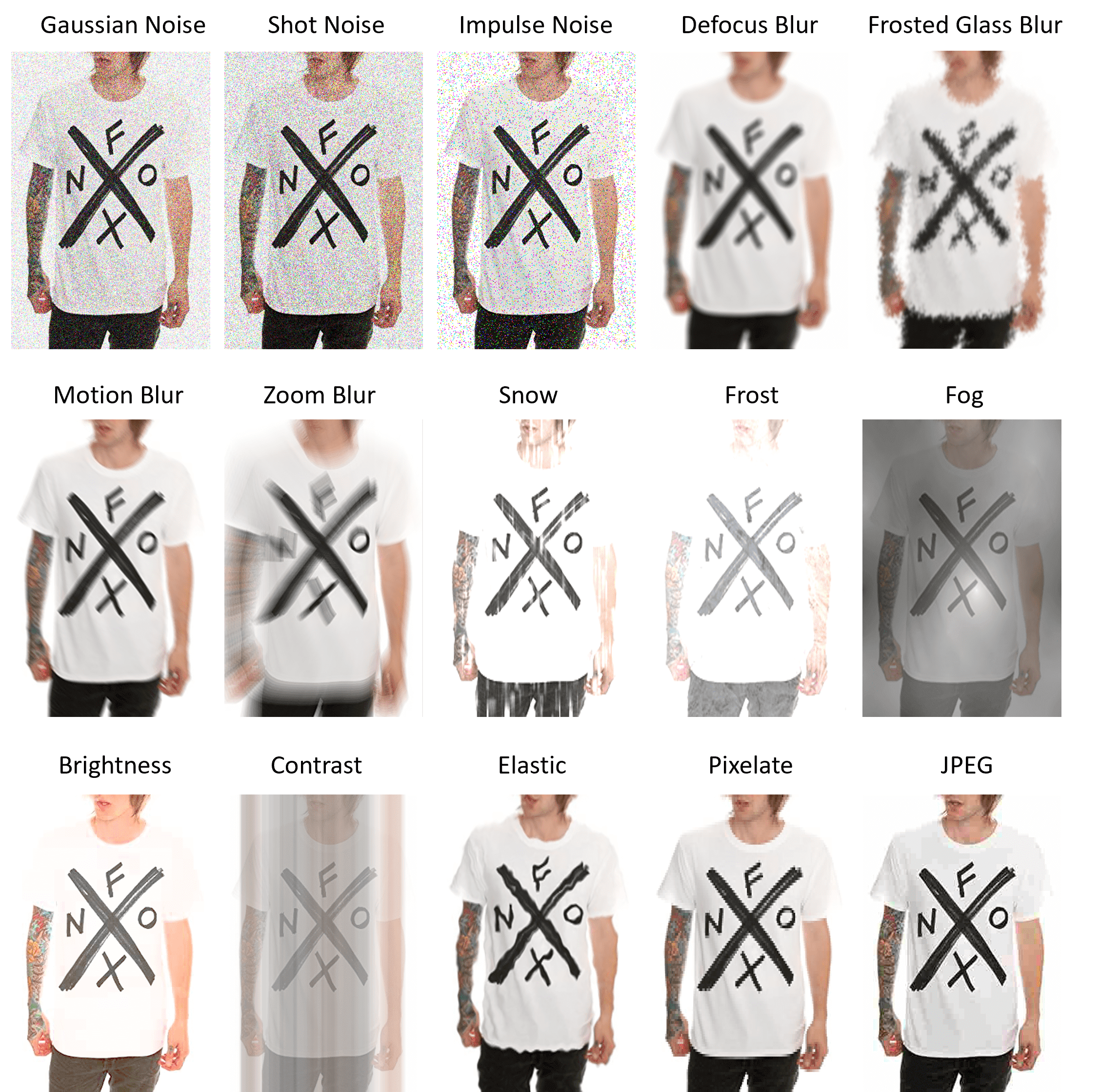}
		(a) Sample visualization with 15 standard image corruptions.
	\end{minipage}
	\begin{minipage}[t]{0.85\textwidth}
		\centering
		\includegraphics[width=\textwidth]{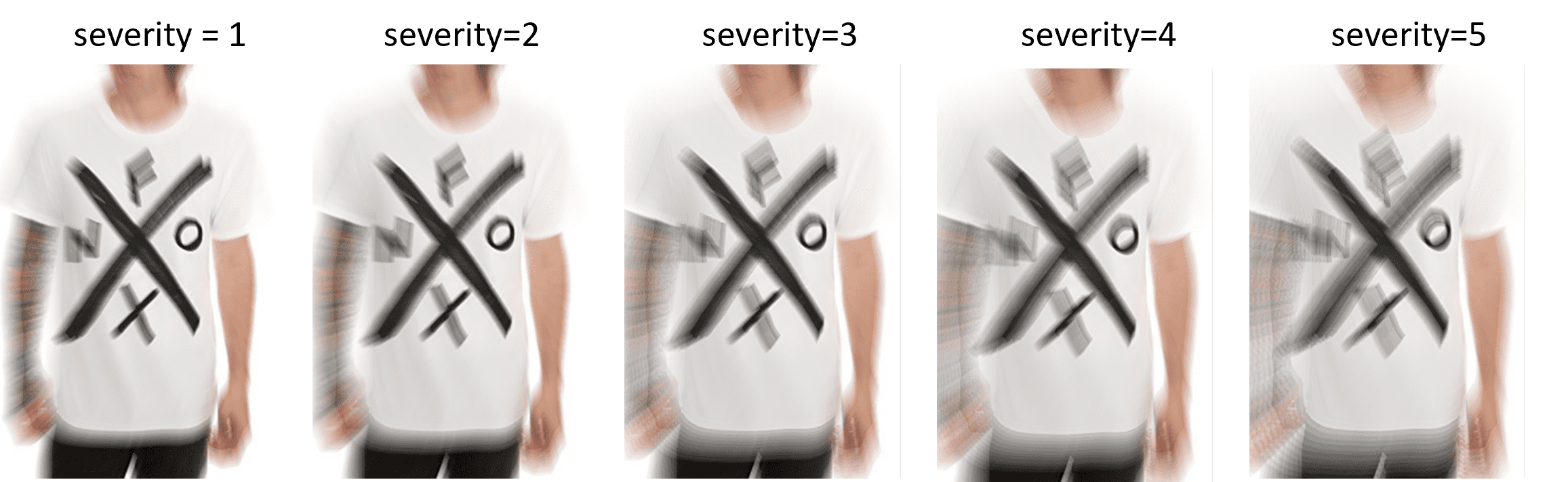}
		(b) Sample visualization of zoom blur corruption with 5 severities.
	\end{minipage}
	\begin{minipage}[t]{0.85\textwidth}
		\centering
		\includegraphics[width=\textwidth]{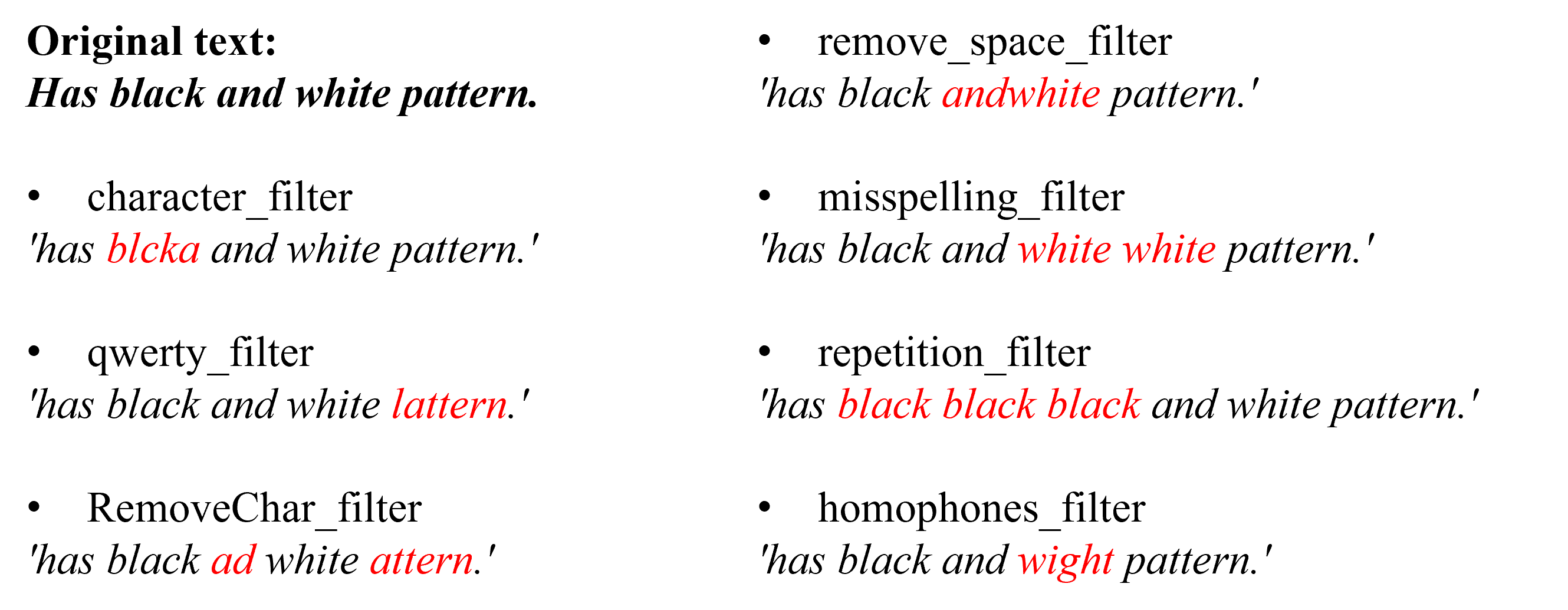}
		(c) Sample visualization of 7 text corruptions.
	\end{minipage}

	\caption{fashionIQ-C sample visualization: (a) 15 standard image corruptions, (b) 5 severities of zoom blur corruption and (c) 7 text corruptions.}
	\label{fig:fashionIQ-C}
\end{figure}

FashionIQ-C follows the same natural corruption in both image and text as in CIRR-C.
We show the visualization samples from FashionIQ-C in Figure.~\ref{fig:fashionIQ-C}.
FashionIQ-C is based on the FashionIQ dataset and implemented with both image corruptions and text corruptions.
We apply 15 standard natural image corruptions, as depicted in Figure~\ref{fig:fashionIQ-C} (a), and demonstrate 5 levels of severity using zoom blur corruption as an example in Figure~\ref{fig:fashionIQ-C} (b).
We further visualize 7 text corruptions in Figure~\ref{fig:fashionIQ-C} (c).

\subsection{Textual corruption definition}
\sst{In this work, we implement 7 natural textual corruptions following~\cite{rychalska2019models}. The definition of the textual corruptions are as follows:
	\begin{itemize}
		\item Swap: Randomly shuffles two characters within a word.
		\item Qwerty: Simulates errors made while writing on a QWERTY-type keyboard. Characters are swapped for their neighbors on the keyboard
		\item RemoveChar: Randomly removes characters from words.
		\item RemoveSpace: Removes a space from text, merging two words.
		\item Misspelling: Misspells words appearing in the Wikipedia list of \href{https://en.wikipedia.org/wiki/Commonly_misspelled_English_words}{commonly misspelled English words}.
		\item Repetition: Randomly repeat words.
		\item Homophone: Changes words into their homophones from the Wikipedia list of \href{https://en.wikipedia.org/wiki/Wikipedia:Lists_of_common_misspellings/Homophones}{common homophones}. The list contains around 500 pairs or triples of homophonic words.
	\end{itemize}
	Examples are shown in Figure.~\ref{fig:CIRR-C} for CIRR-C and Figure.~\ref{fig:fashionIQ-C} for FashionIQ-C respectively.}

\subsection{CIRR-D visualization}
\begin{figure}[htbp]
	\centering
	\begin{minipage}[t]{0.9\textwidth}
		\centering
		\includegraphics[width=\textwidth]{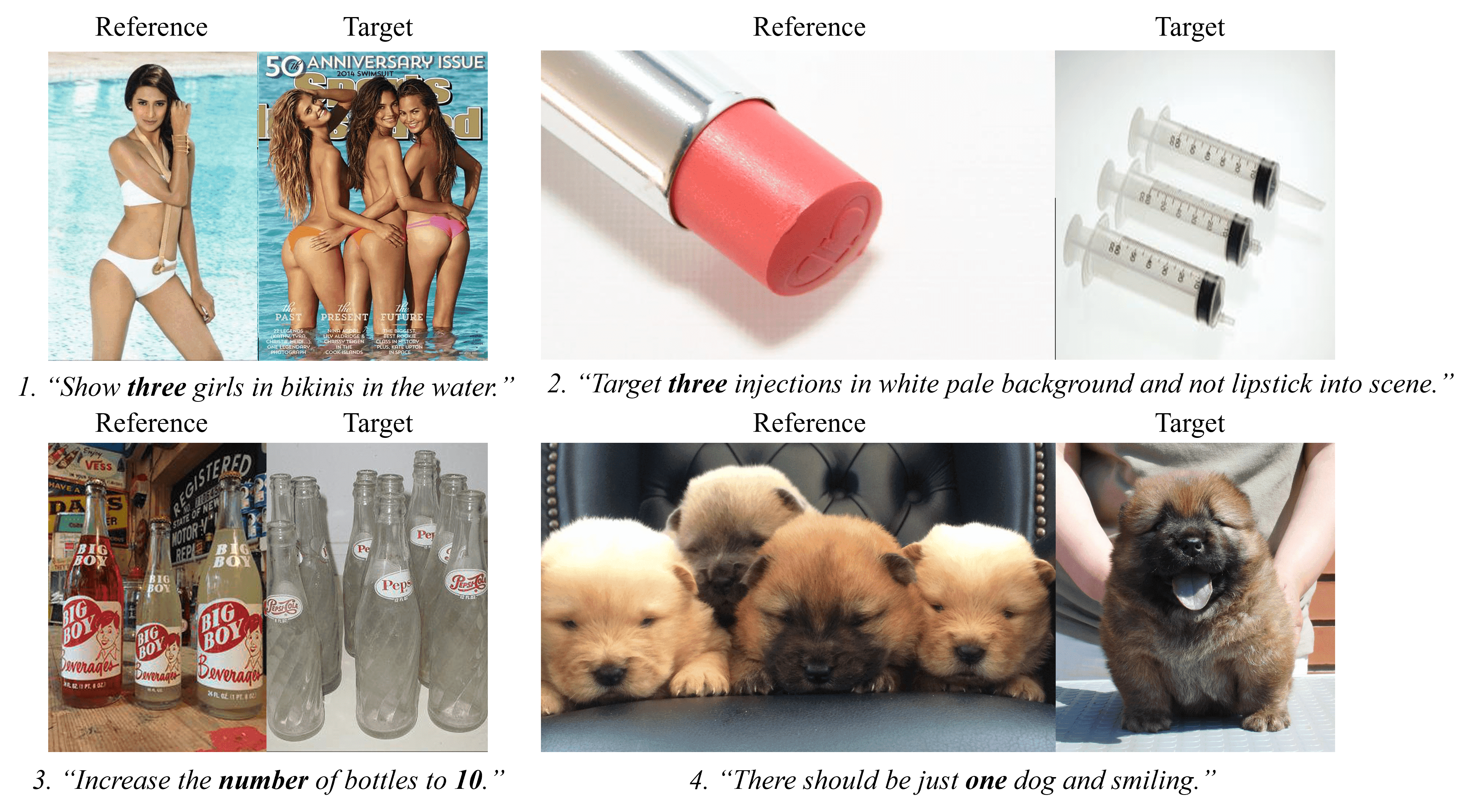}
		(a) Numerical queries from original CIRR dataset, four triplets are included.
	\end{minipage}
	\begin{minipage}[t]{0.9\textwidth}
		\centering
		\includegraphics[width=\textwidth]{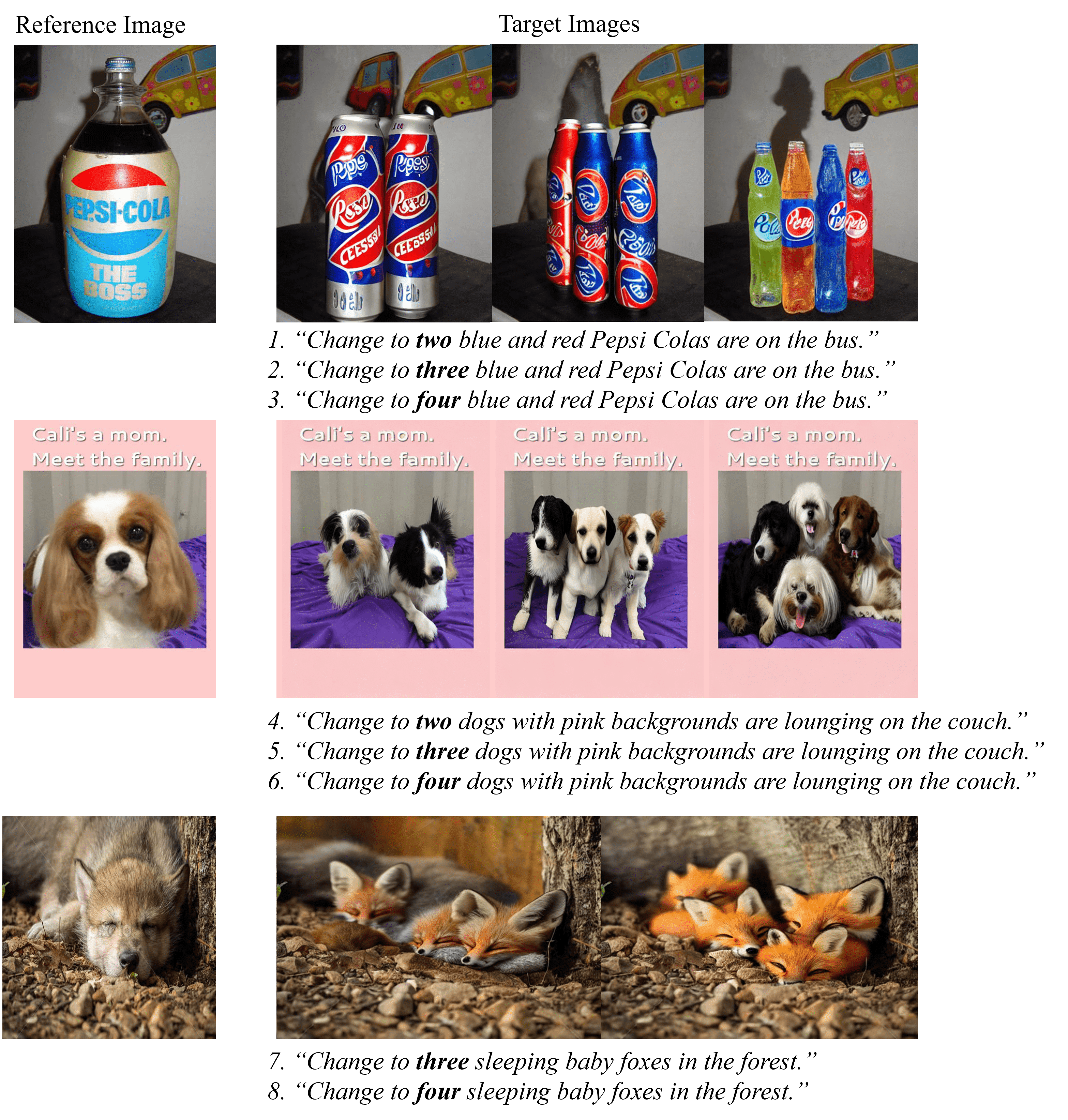}
		(b) Our generated numerical queries, eight triplets are included.
	\end{minipage}
	\caption{CIRR-D sample visualization for numerical queries.}
	\label{fig:numerical}
\end{figure}

\begin{figure}[htbp]
	\centering
	\begin{minipage}[t]{1.0\textwidth}
		\centering
		\includegraphics[width=\textwidth]{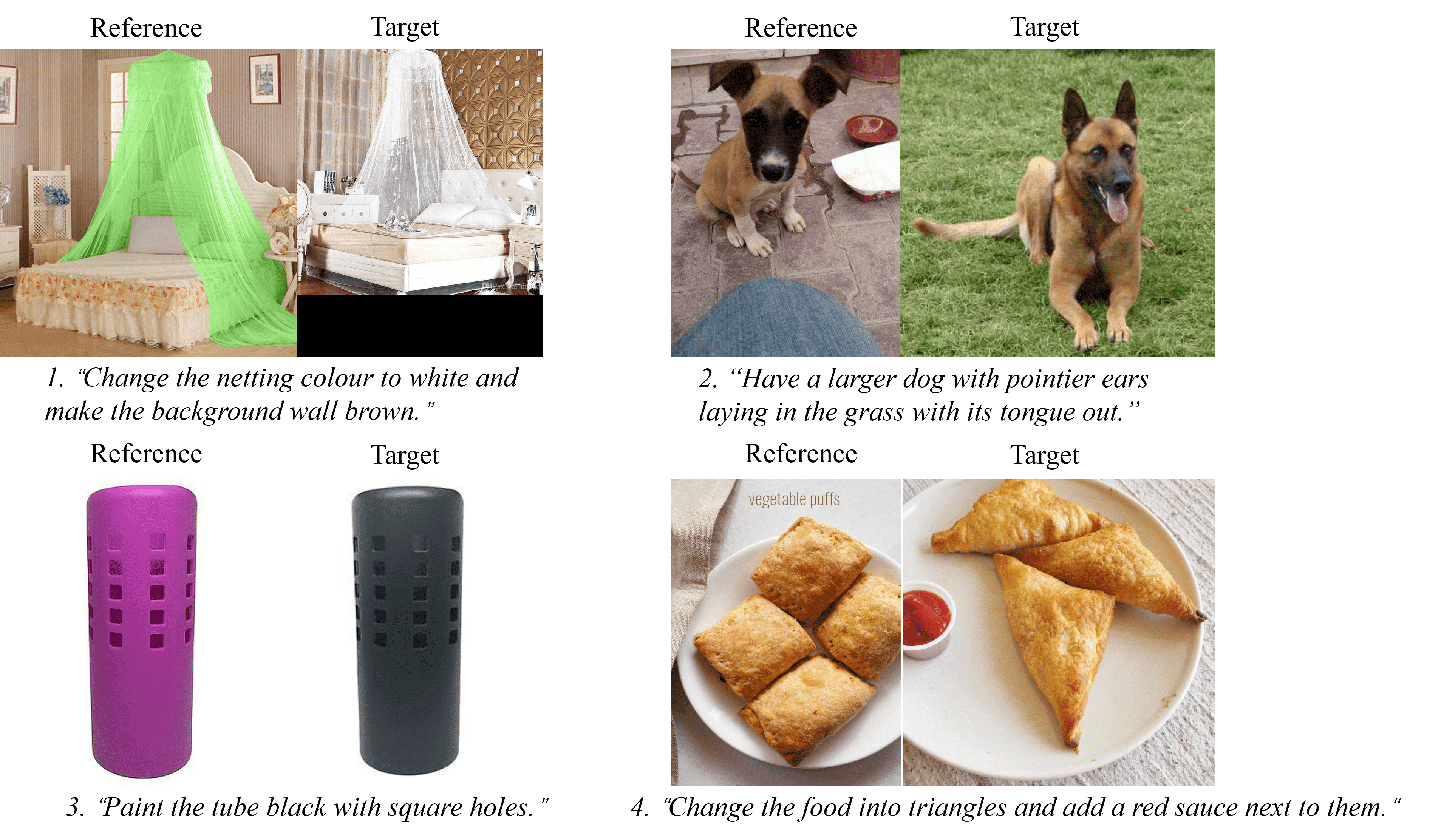}
		(a) Attribute queries from original CIRR dataset, four triplets are included.
	\end{minipage}
	\begin{minipage}[t]{1.0\textwidth}
		\centering
		\includegraphics[width=\textwidth]{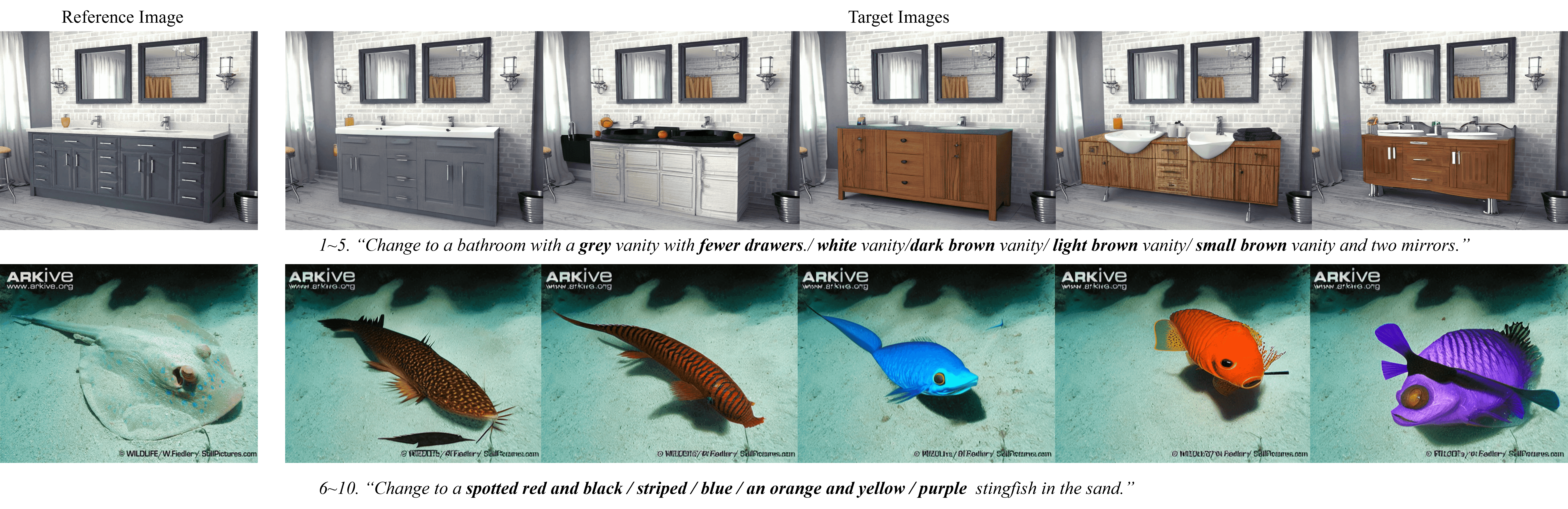}
		(b) Our generated attribute queries, 10 triplets are included.
	\end{minipage}
	\caption{CIRR-D sample visualization for attribute queries including color, shape and size.}
	\label{fig:attribute}
\end{figure}

\begin{figure}[htbp]
	\centering
	\begin{minipage}[t]{0.9\textwidth}
		\centering
		\includegraphics[width=\textwidth]{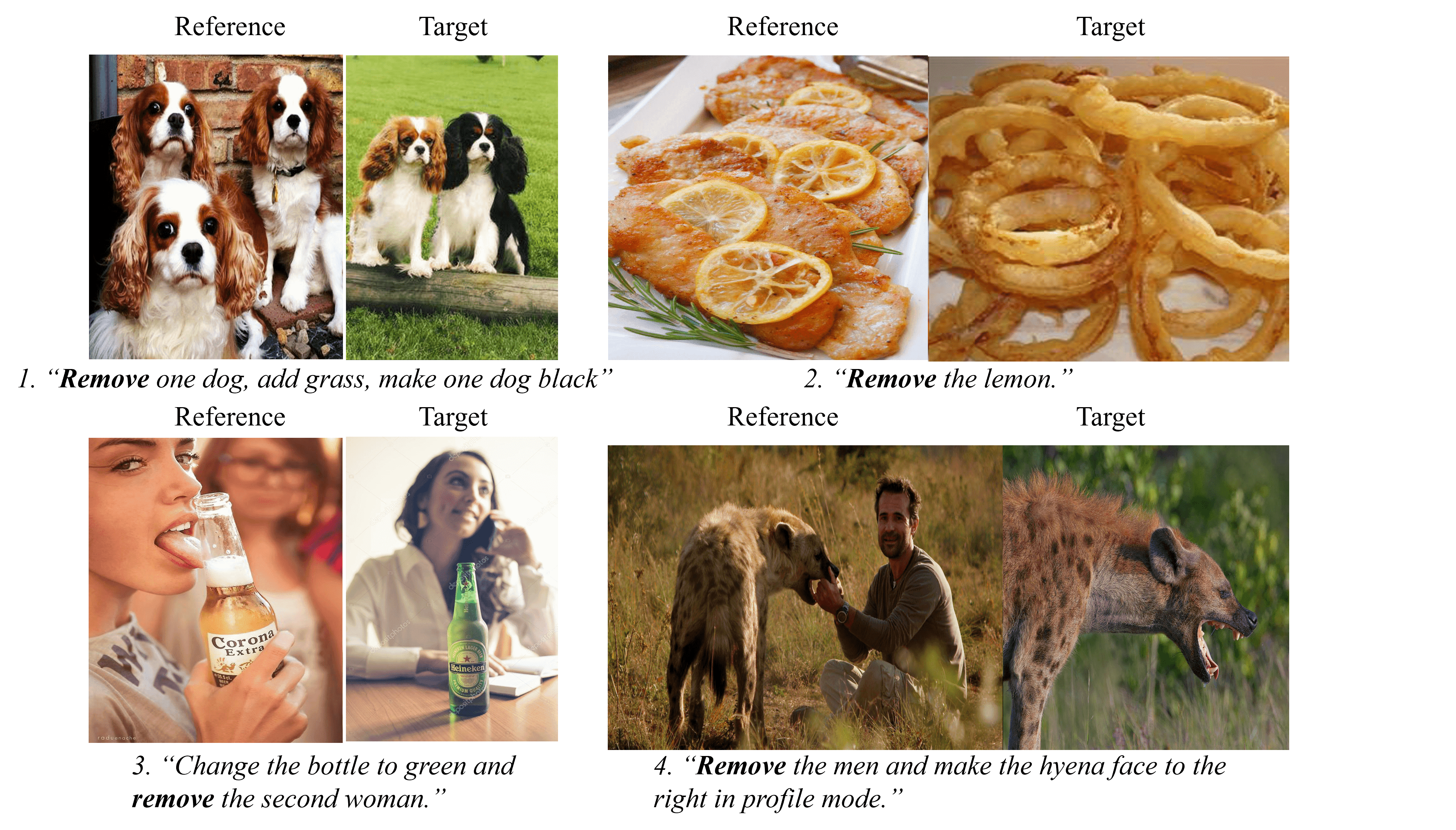}
		(a) Object removal queries from original CIRR dataset, four triplets are included.
	\end{minipage}
	\begin{minipage}[t]{0.9\textwidth}
		\centering
		\includegraphics[width=\textwidth]{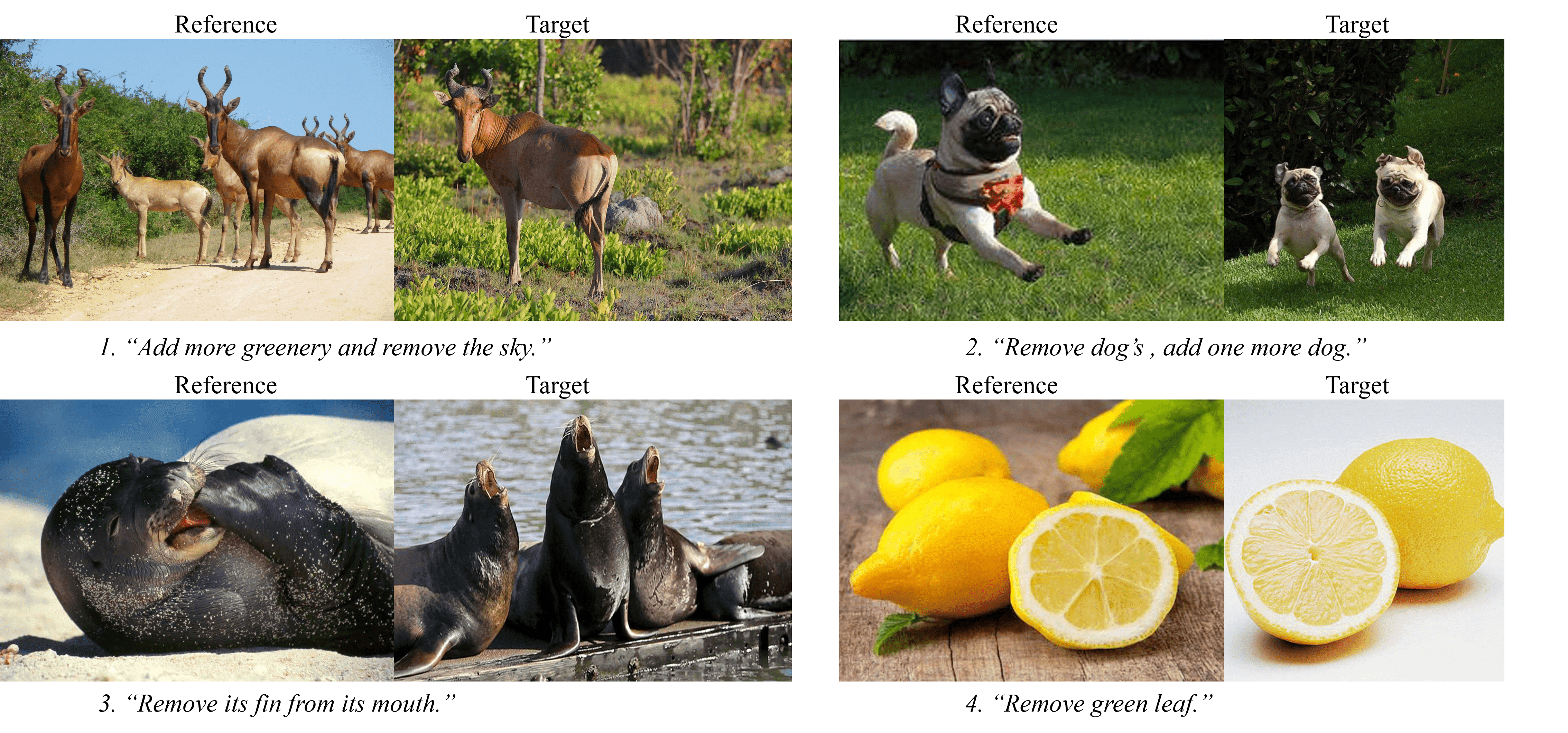}
		(b) Object removal queries from extend captions of original CIRR dataset, four triplets are included.
	\end{minipage}
	\begin{minipage}[t]{0.8\textwidth}
		\centering
		\includegraphics[width=\textwidth]{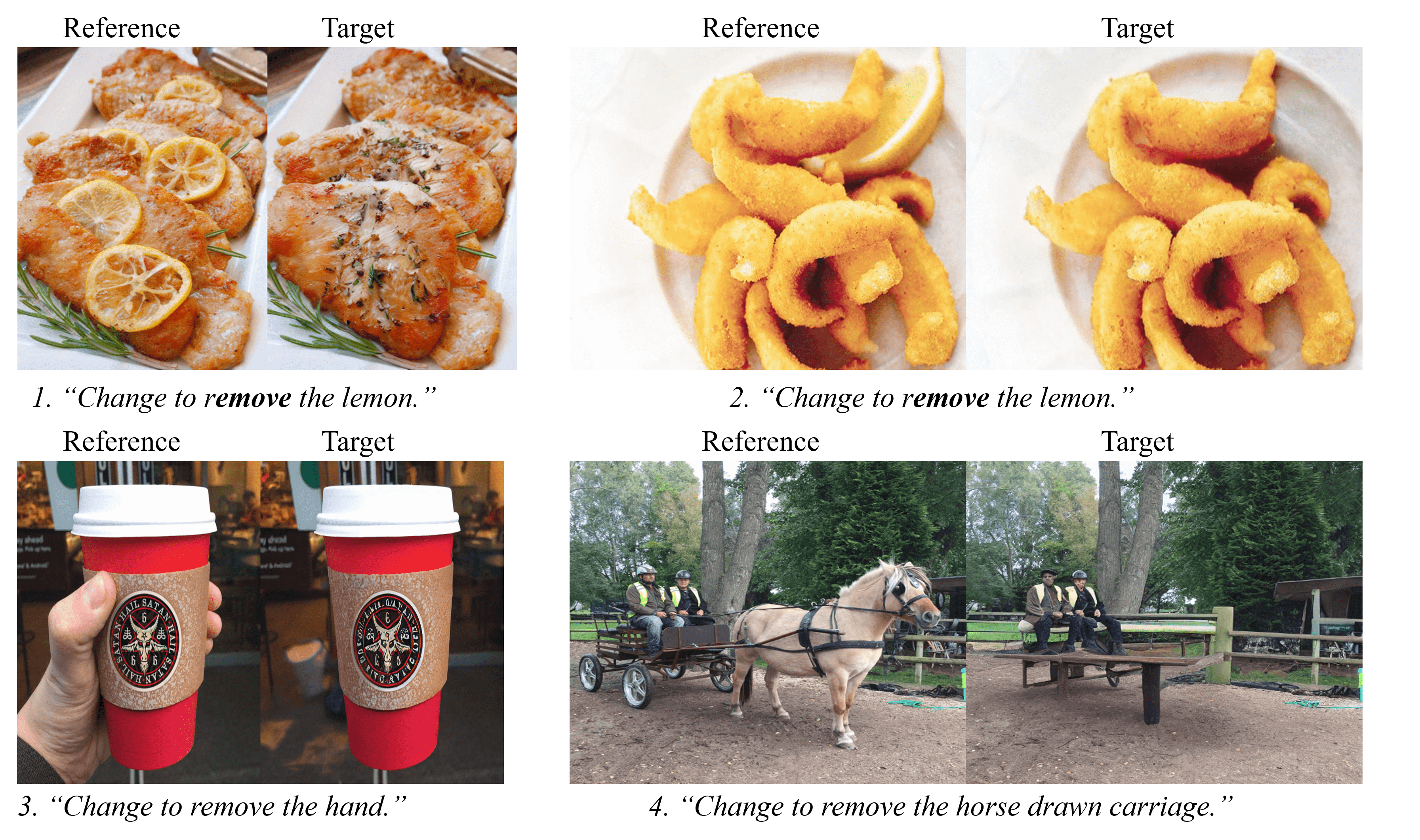}
		(c) Our generated object removal queries, four triplets are included.
	\end{minipage}
	\caption{CIRR-D sample visualization for object removal queries.}
	\label{fig:removal}
\end{figure}

\begin{figure}[htbp]
	\centering
	\begin{minipage}[t]{1.0\textwidth}
		\centering
		\includegraphics[width=\textwidth]{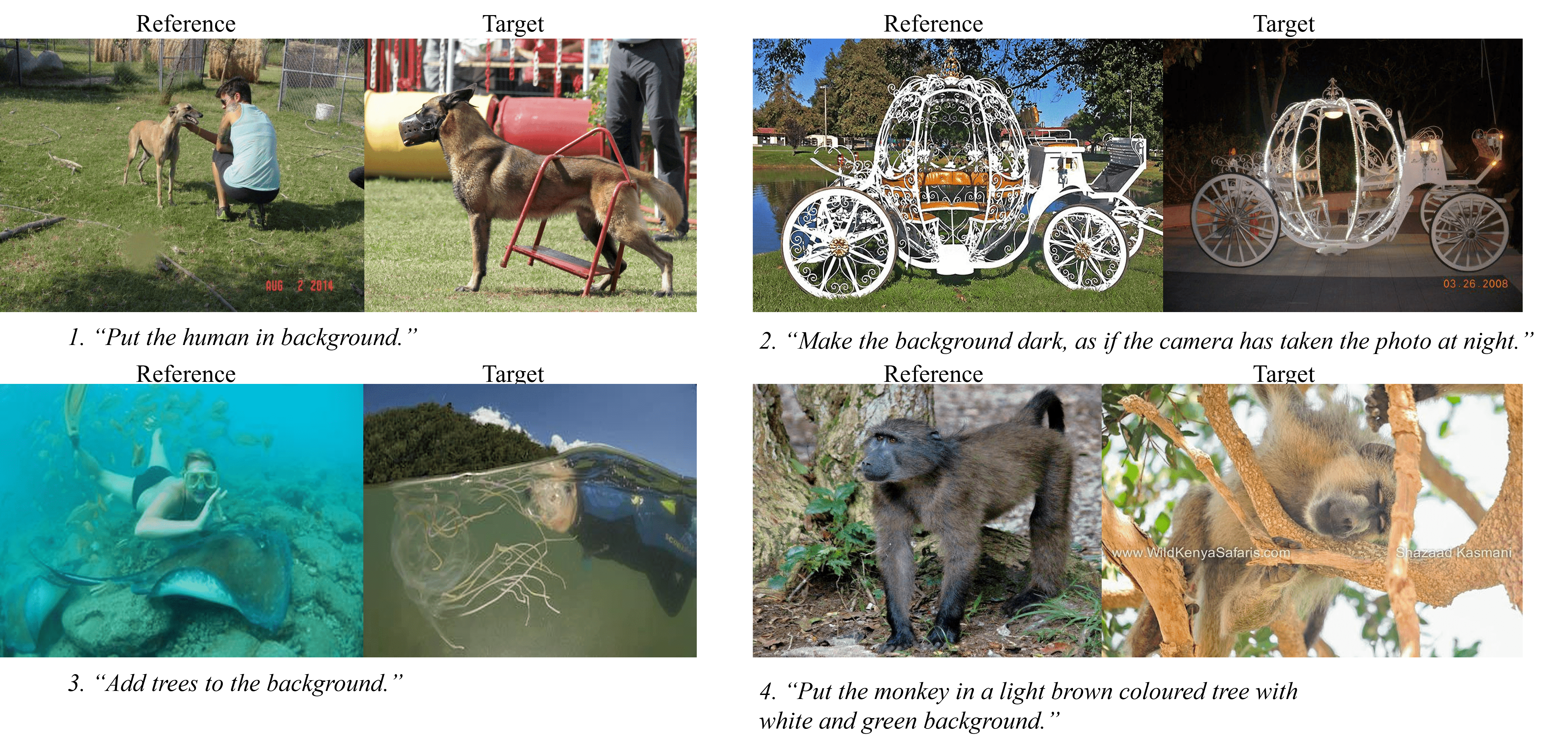}
		(a) Background queries from original CIRR dataset, four triplets are included.
	\end{minipage}
	\begin{minipage}[t]{1.0\textwidth}
		\centering
		\includegraphics[width=\textwidth]{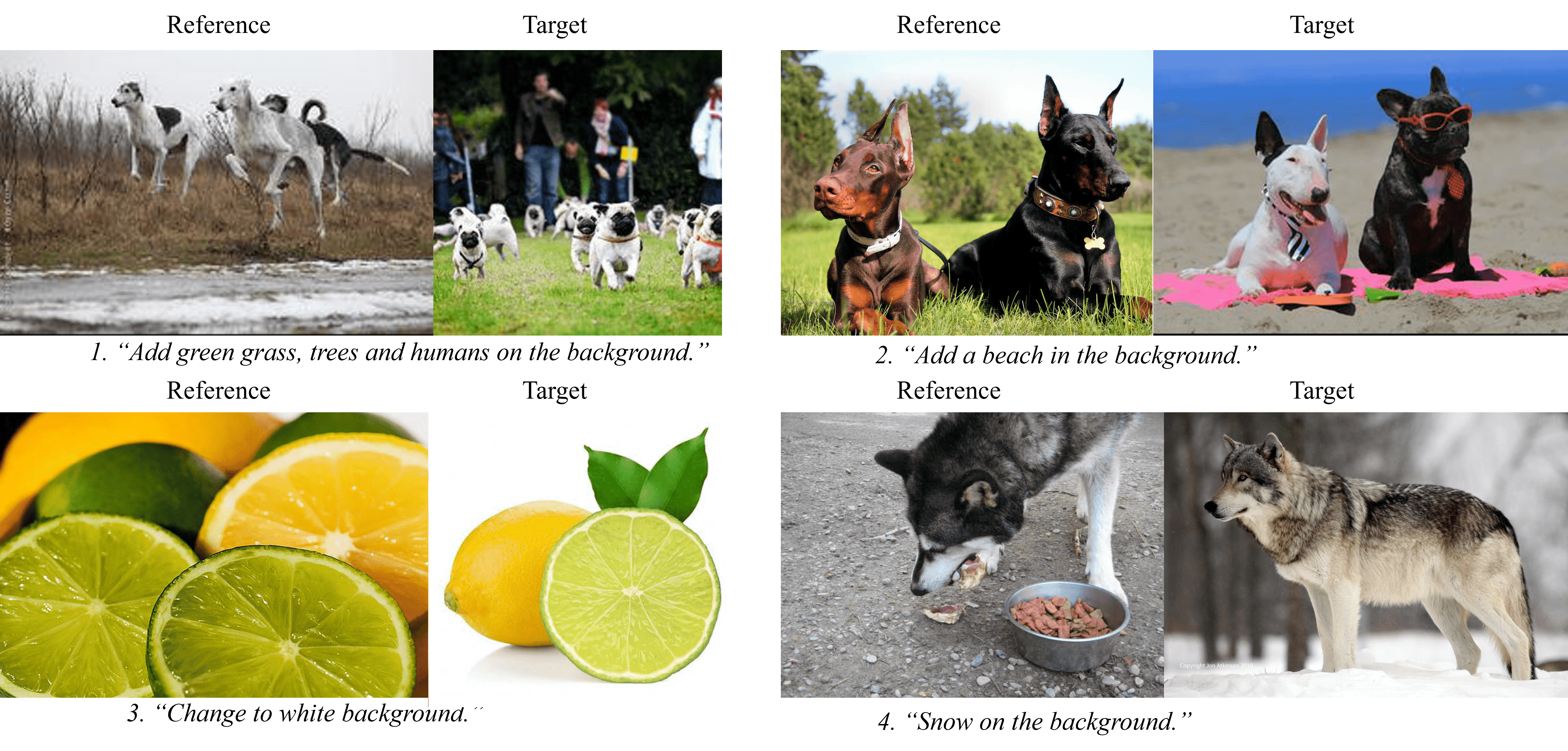}
		(b) Background variation queries from extend captions of original CIRR dataset, four triplets are included.
	\end{minipage}
	\caption{CIRR-D sample visualization for background variations.}
	\label{fig:background}
\end{figure}

\begin{figure}[t]
	\centering
	\includegraphics[width=1.0\textwidth]{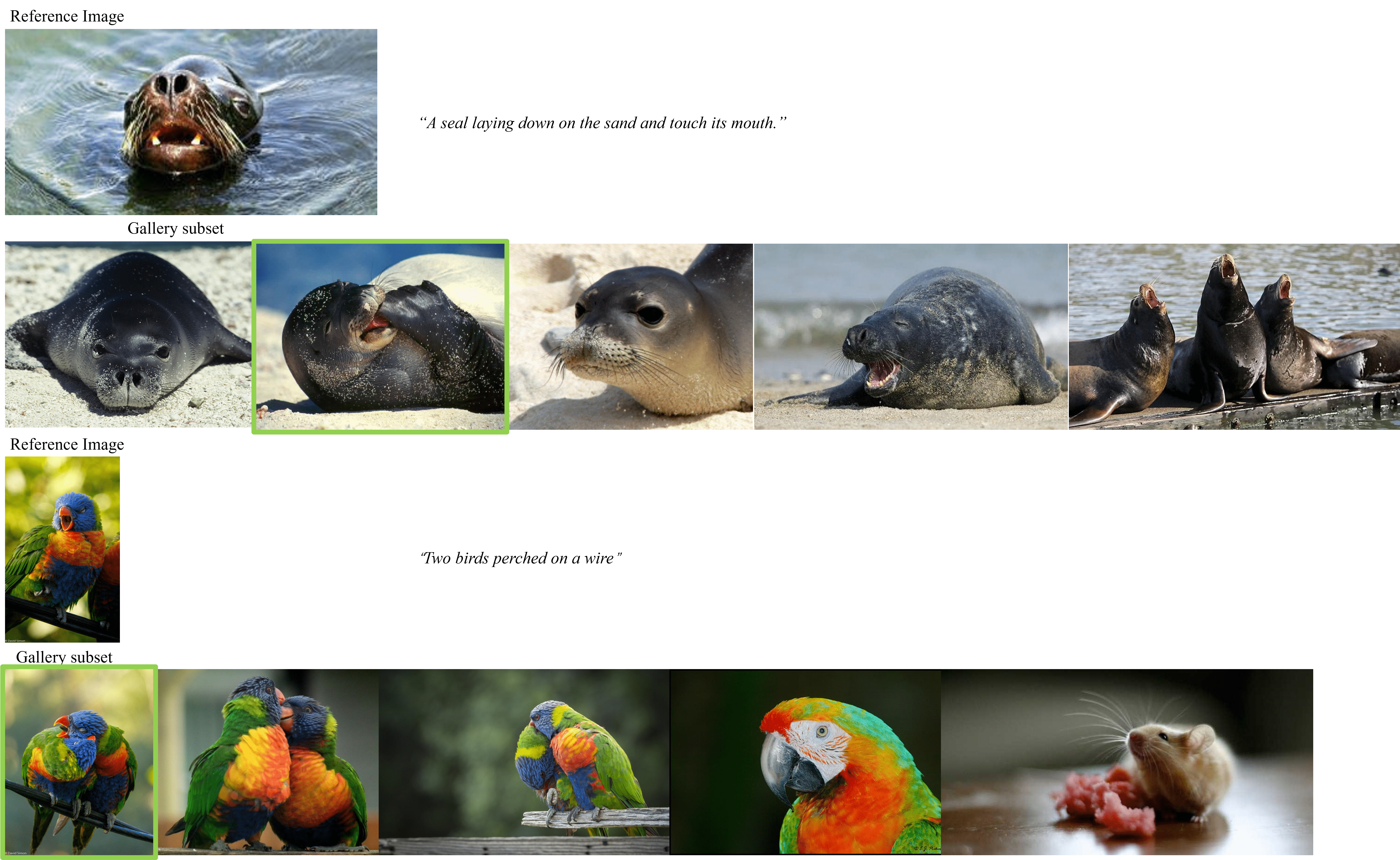}
	\caption{CIRR-D sample visualization for fine-grained variation queries, 2 triplets are included.
		The images with a green border are the correct targets, while the other images are highly similar composing the gallery set.}
	\label{fig:fine_grained}
\end{figure}

To detect textual understanding ability, we build a CIRR-D dataset with five different types of queries containing specific instructions to probe five different abilities
The source of the CIRR-D dataset is from the original CIRR, CIRR extends caption and our generated synthetic images.
The triplets from the original CIRR dataset are normally with obvious variations while the synthetic triplets are normally following the same structure and only local variations.
Some extended caption from the original CIRR dataset can only supply partial difference and cannot locate the target images.
Therefore, we manually remove samples and retain only those triplets where the extended caption can provide sufficient variations.
In detail, we visualize numerical samples in Figure~\ref{fig:numerical}, which is composed of triplets from the original CIRR dataset in Figure~\ref{fig:numerical} (a) and our generated synthetic triplets in Figure.\ref{fig:numerical} (b).
Attribute variation visualization samples are shown in Figure~\ref{fig:attribute}, which is composed of triplets from the original CIRR dataset in Figure~\ref{fig:attribute} (a) and our generated synthetic triplets in Figure.~\ref{fig:attribute} (b).
To evaluate object removal ability, the triplet source consists of three aspects.
We visualize object removal triplets from the original CIRR dataset in Figure.~\ref{fig:removal} (a), extended caption triplets in Figure.~\ref{fig:removal} (b) and our generated synthetic triplets in Figure.~\ref{fig:removal} (c).
we visualize background variations samples in Figure~\ref{fig:background}, which is composed of triplets from original CIRR dataset in Figure~\ref{fig:background} (a) and from extended captions in Figure.\ref{fig:background} (b).
The evaluation of fine-grained variations follows the original CIRR dataset, whose gallery set is composed of 5 highly similar images as shown in Figure.~\ref{fig:fine_grained}.

\begin{table}[t!]
	\centering
	\small
	\caption{Details of CIRR-D dataset. The first column is the number of images. The rest columns contain the number of triplets for five probing abilities.}
	\begin{tabular}{l|c | c c c c c}
		\toprule
		               & Images & Numerical & Attribute & Removal & Background & Fine-grained \\  
		\midrule

		Val.           & $2297$ & $820$     & $1397$    & $233$   & $358$      & $4181$       \\  
		Extend caption & -      & -         & -         & $505$   & $812$      & -            \\  
		Synthetic      & $1245$ & $305$     & $700$     & $140$   & -          & -            \\ 
		\midrule
		Total          & $3542$ & $1125$    & $2097$    & $878$   & $1170$     & $4181$       \\ 


		\bottomrule
	\end{tabular}  \vspace{-0.3cm}
	\label{tab:synthetic}
\end{table}

\begin{table}[t!]
	\centering
	\small
	\caption{Details of modality fusion for the evaluated models in this study. $R_i,M_t,T_i$ represent reference image, modified text and target image feature respectively. $C$ donates the composed of the two features. <,> represents cosine similarity. }
	{\begin{tabular}{l|c|c|c|c}
			\toprule
			Model                                 & $C_{RiMt}$            & $C_{RiTi}$  & $C_{MtTi}$  & distance                                     \\
			\midrule
			TIRG~\cite{vo2019composing}           & cat+residual          & -           & -           & <$C_{RiMt}$, $T_i$>                          \\
			MAAF~\cite{dodds2020modality}         & self attn+ cross attn & -           & -           & <$C_{RiMt}$, $T_i$>                          \\
			Artemis~\cite{delmas2022artemis}      & dot product           & dot product & dot product & <$C_{RiMt}$,$C_{TiMt}$> + <$C_{TiMt}$,$M_t$> \\
			CIRPLANT~\cite{liu2021image}          & transformer           & -           & -           & <$C_{RiMt}$, $T_i$>                          \\
			CLIP4CIR~\cite{baldrati2022effective} & cat + residual        & -           & -           & <$C_{RiMt}$, $T_i$>                          \\
			FashionViL~\cite{han2022fashionvil}   & transformer           & -           & -           & <$C_{RiMt}$, $T_i$>                          \\
			\bottomrule
		\end{tabular}  \vspace{-0.3cm}}
	\label{tab:models_detail}
\end{table}

\begin{table}[t!]
	\centering
	\small
	\caption{Relative robustness score for text-image composed retrieval under 7 natural text corruptions in CIRR-C recall@10 and FashionIQ-C recall@10 on average of three categories. \sst{Recall@10 performance under clean conditions on the left. }\textbf{Bold} are the highest relative robustness among the five compared methods.}
	{\setlength\tabcolsep{6pt}
	\begin{tabular}{l|c|c c c c| c c c}
		\multicolumn{1}{c}{}     & \multicolumn{1}{c}{}        & \multicolumn{4}{c}{Character} & \multicolumn{3}{c}{Word}                                                                                                                                    \\
		\toprule
		\textbf{CIRR-C}          & \scriptsize{Clean}          & \scriptsize{Swap}             & \scriptsize{QWERTY}      & \scriptsize{RemoveChar} & \scriptsize{RemoveSpace} & \scriptsize{Misspelling} & \scriptsize{Repetition} & \scriptsize{Homophone} \\
		\midrule
		Text-only                     &    \sst{51.2}     & $0.75$                        & $ 0.74$                  & $0.78$                  & $1.0$                    & $0.99$                   & $0.98$                  & $0.92$                 \\
		\midrule

		TIRG~\cite{vo2019composing}     &    \sst{55.1}   & $0.77$                        & $0.76$                   & $0.80$                  & $1.0$                    & $0.98$                   & $1.0$                   & $0.89$                 \\

		MAAF~\cite{dodds2020modality}     &  \sst{49.9}   & $\textbf{0.95}$               & $\textbf{0.97}$          & $\textbf{0.96}$         & $\textbf{1.0}$           & $\textbf{1.0}$           & $\textbf{1.0}$          & $\textbf{0.97}$        \\
		ARTEMIS~\cite{delmas2022artemis}    &  \sst{59.0} & $0.61$                        & $0.58$                   & $0.65$                  & $\textbf{1.0}$           & $0.98$                   & $0.98$                  & $0.82$                 \\

		CIRPLANT~\cite{liu2021image}         & \sst{68.8}  & $0.92$                        & $0.93$                   & $0.93$                  & $\textbf{1.0}$           & $\textbf{1.0}$           & $\textbf{1.0}$          & $\textbf{0.97}$        \\
		CLIP4CIR~\cite{baldrati2022effective} & \sst{\textbf{80.3}}  & $0.89$                        & $ 0.89$                  & $0.90$                  & $\textbf{1.0}$           & $\textbf{1.0}$           & $0.99$                  & $\textbf{0.97}$        \\
		\bottomrule
	\end{tabular}
	\label{tab:robustness_text}
	\begin{tabular}{l|c|c c c c| c c c}
		\multicolumn{1}{c}{}             & \multicolumn{1}{c}{}        & \multicolumn{4}{c}{Character} & \multicolumn{3}{c}{Word}                                                                                                                                    \\
		\toprule
		\textbf{FashionIQ-C}               & \scriptsize{Clean}       & \scriptsize{Swap}             & \scriptsize{QWERTY}      & \scriptsize{RemoveChar} & \scriptsize{RemoveSpace} & \scriptsize{Misspelling} & \scriptsize{Repetition} & \scriptsize{Homophone} \\
		\midrule
		TIRG~\cite{vo2019composing}         &   \sst{23.8}   & $0.26$                        & $ 0.20$                 & $0.29$                  & $ 0.66$                  & $ 0.63$                  & $ 0.61$                 & $ 0.52$                \\

		MAAF~\cite{dodds2020modality}    &   \sst{23.4}      & $0.40$                        & $0.39$                   & $0.39$                  & $0.70$                   & $0.68$                   & $0.68$                  & $0.62$                 \\
		ARTEMIS~\cite{delmas2022artemis}      &  \sst{24.9}  & $0.25$                        & $0.20$                   & $0.31$                  & $0.70$                   & $0.67$                   & $0.67$                  & $0.55$                 \\

		FashionViL~\cite{han2022fashionvil}   &  \sst{23.4}  & $\textbf{0.55}$               & $\textbf{0.59}$          & $\textbf{0.60}$         & $\textbf{0.86}$          & $\textbf{0.84}$          & $\textbf{0.85}$         & $\textbf{0.76}$        \\
		CLIP4CIR~\cite{baldrati2022effective} &  \sst{\textbf{35.9}}  & $0.52$                        & $0.51$                   & $0.54$                  & $0.71$                   & $0.70$                   & $0.69$                  & $0.67$                 \\

		\bottomrule
	\end{tabular} }
\end{table}

\section{More experiments result}

\subsection{Textual robustness against natural corruption}
\label{supp:textual}
Comparing the relative robustness against textual corruption in Table~\ref{tab:robustness_text} and visual corruption in Table~\ref{tab:robustness_image}, we can observe that the robustness is higher against textual corruptions.
Among the compared models, MAAF and FashionViL show the highest robustness in open domain and fashion domain respectively.
This aligns with the findings we discovered regarding vision corruptions, where large pretrained models (FashionViL with fashion-specific pretraining) and models with sufficient modality fusion result to higher robustness.
Additionally, comparing CLIP4CIR and the Text-only retrieval method implementing CLIP text embedding, we find out the robustness can be boosted after fusion with vision modality.
However, with images corrupted, CLIP4CIR shows lower robustness than text-only model, from which \emph{we speculate that aligned clean image feature can boost the robustness, while the corrupted image feature will impair robustness.}

\subsection{Fine grained subset analysis of CIRR-C}

\begin{figure}[t]
	\centering
	\includegraphics[width=1.0\textwidth]{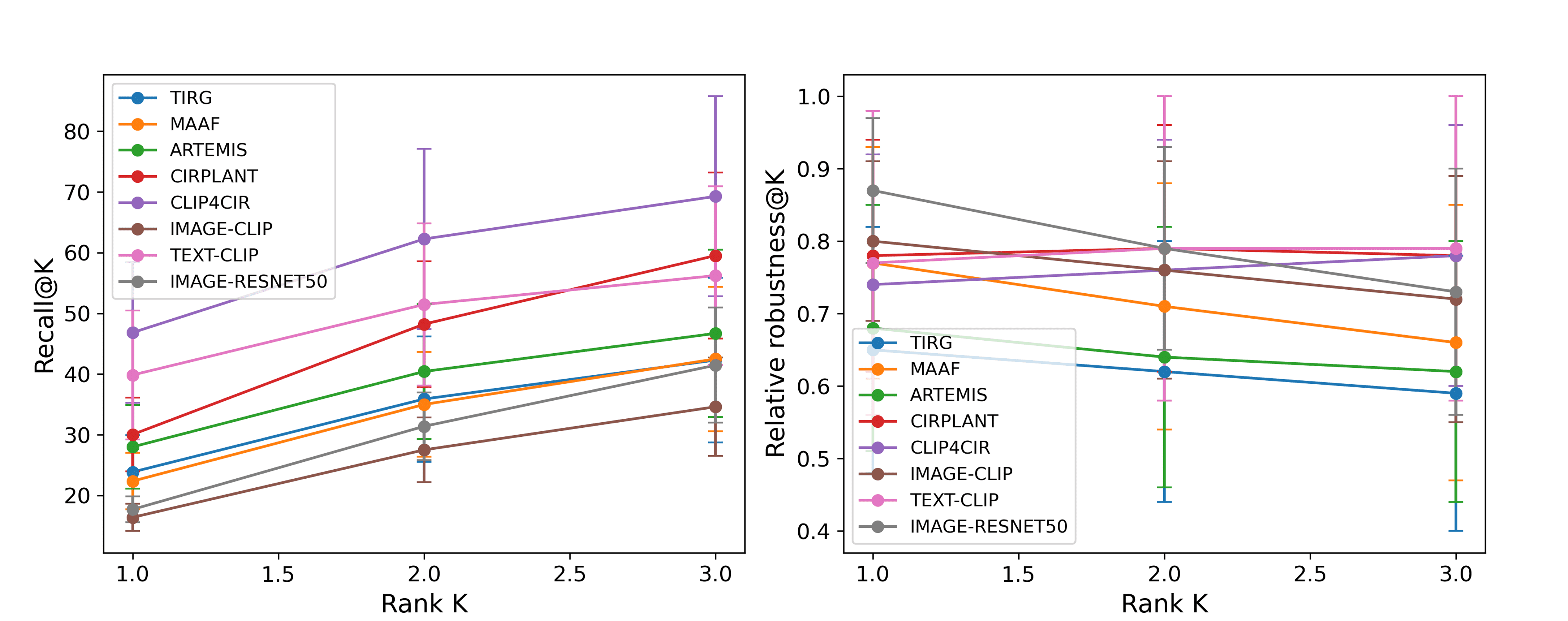}
	\caption{Models average performance in CIRR subset under 15 vision corruptions. Left: Recall vs. rank K.
		Right: Relative robusentess vs. rank K. }
	\label{fig:subset}
\end{figure}

In this section, we supplement some experimental results.
As shown in Figure.~\ref{fig:subset}, we visualize the recall performance on the CIRR-C subset.
Comparing the subset retrieval with the whole gallery in CIRR-C (shown in the main paper Figure.\ref{fig:accuracies_compare}), we can observe that subset relative robustness (range from 0.6 to 0.9) overall is higher than the whole set (range from 0.4 to 0.8).
This result suggests that a smaller gallery can lead to more stable retrieval.
In essence, the overall trend aligns with retrieval on all images: CLIP4CIR consistently performs the best, while IMAGE-ONLY with CLIP embedding consistently exhibits the worst retrieval performance.

\subsection{Subcategories analysis of FashionIQ-C}

For detailed results in the FashionIQ-C dataset, we report the results on the three categories, namely dress, shirt and toptee respectively, as shown in Table.~\ref{tab:fashion_recall10_cat}.
Overall, a similar trend is observed across the three categories, with FashionViL and CLIP4CIR consistently exhibiting the highest relative robustness.
In the shirt category, overall robustness tends to be slightly higher than in the dress and toptee categories.
We further report the recall@10 performance on FashionIQ-C dataset as shown in Table.~\ref{tab:fashion_recall10_acc}.
By comparing the relative robustness in Table.~\ref{tab:fashion_recall10_cat} and corresponding recall performance in Table.~\ref{tab:fashion_recall10_acc}, we can find out higher robustness doesn't mean higher recall performance.
As according to the definition of relative robustness: $\gamma=1-\left(R_c-R_p\right) / R_c$ following ~\cite{hendrycks2019robustness}, lower recall performance under clean condition $R_c$ will lead to higher relative robustness $\gamma$.

\begin{table}[t!]
	\centering
	\small
	\caption{Relative robustness score for text-image composed retrieval under 15 natural image corruptions in FashionIQ-C Recall@10 for dress, shirt and toptee respectively. \textbf{Bold} is the highest relative robustness for the five composed retrieval methods.}
	{\setlength\tabcolsep{1.5pt}
		\begin{tabular}{l|c c c|c c c c|c c c c|c c c c}
			\multicolumn{1}{c}{}                  & \multicolumn{3}{c}{Noise} & \multicolumn{4}{c}{Blur} & \multicolumn{4}{c}{Weather} & \multicolumn{4}{c}{Digital}                                                                                                                                                                                                                                          \\

			\toprule
			\textbf{FashionIQ-C Dress}            & \scriptsize{Gauss.}       & \scriptsize{Shot}        & \scriptsize{Impluse}        & \scriptsize{Defocus}        & \scriptsize{Glass} & \scriptsize{Motion} & \scriptsize{Zoom} & \scriptsize{Snow} & \scriptsize{Frost} & \scriptsize{Fog} & \scriptsize{Bright} & \scriptsize{Contrast} & \scriptsize{Elastic} & \scriptsize{Pixel} & \scriptsize{JPEG} \\
			\midrule



			TIRG~\cite{vo2019composing}           & 0.21                      & 0.18                     & 0.17                        & 0.37                        & 0.22               & 0.64                & 0.58              & 0.33              & 0.25               & 0.35             & 0.63                & 0.12                  & 0.62                 & 0.82               & 0.85              \\
			MAAF~\cite{dodds2020modality}         & 0.30                      & 0.24                     & 0.22                        & 0.42                        & 0.19               & 0.65                & 0.56              & 0.28              & 0.21               & 0.32             & 0.58                & 0.10                  & 0.54                 & 0.78               & 0.81              \\
			ARTEMIS~\cite{delmas2022artemis}      & 0.23                      & 0.22                     & 0.18                        & 0.38                        & 0.24               & 0.66                & 0.62              & 0.39              & 0.26               & 0.37             & 0.59                & 0.14                  & 0.67                 & 0.85               & 0.9               \\

			FashionViL~\cite{han2022fashionvil}   & 0.21                      & 0.22                     & 0.23                        & 0.38                        & \textbf{0.34}      & \textbf{0.84}       & \textbf{ 0.72}    & 0.29              & 0.29               & 0.3              & \textbf{0.79}       & 0.13                  & \textbf{0.88}        & \textbf{ 1.1}      & \textbf{ 1.1}     \\
			CLIP4CIR~\cite{baldrati2022effective} & \textbf{0.44}             & \textbf{0.38}            & \textbf{0.44}               & \textbf{0.54}               & 0.24               & 0.74                & 0.52              & \textbf{0.41}     & \textbf{0.36}      & \textbf{0.55}    & 0.68                & \textbf{0.16}         & 0.42                 & 0.75               & 0.82              \\

			\bottomrule
		\end{tabular}

		\begin{tabular}{l|c c c|c c c c|c c c c|c c c c}
			\multicolumn{1}{c}{}                  & \multicolumn{3}{c}{Noise} & \multicolumn{4}{c}{Blur} & \multicolumn{4}{c}{Weather} & \multicolumn{4}{c}{Digital}                                                                                                                                                                                                                                          \\

			\toprule
			\textbf{FashionIQ-C Shirt}            & \scriptsize{Gauss.}       & \scriptsize{Shot}        & \scriptsize{Impluse}        & \scriptsize{Defocus}        & \scriptsize{Glass} & \scriptsize{Motion} & \scriptsize{Zoom} & \scriptsize{Snow} & \scriptsize{Frost} & \scriptsize{Fog} & \scriptsize{Bright} & \scriptsize{Contrast} & \scriptsize{Elastic} & \scriptsize{Pixel} & \scriptsize{JPEG} \\



			\midrule
			TIRG~\cite{vo2019composing}           & 0.33                      & 0.32                     & 0.27                        & 0.28                        & 0.20               & 0.57                & 0.54              & 0.32              & 0.28               & 0.37             & 0.51                & 0.15                  & 0.60                 & 0.86               & 0.81              \\
			MAAF~\cite{dodds2020modality}         & 0.33                      & 0.30                     & 0.27                        & 0.46                        & 0.20               & 0.67                & 0.50              & 0.30              & 0.27               & 0.34             & 0.47                & 0.16                  & 0.57                 & 0.84               & 0.79              \\
			ARTEMIS~\cite{delmas2022artemis}      & 0.27                      & 0.28                     & 0.25                        & 0.39                        & \textbf{ 0.26 }    & 0.62                & \textbf{ 0.61}    & 0.36              & 0.24               & 0.38             & 0.54                & 0.16                  & 0.61                 & 0.84               & 0.88              \\

			FashionViL~\cite{han2022fashionvil}   & 0.29                      & 0.34                     & 0.26                        & 0.38                        & \textbf{0.26 }     & \textbf{0.77 }      & 0.6               & 0.33              & 0.32               & 0.37             & 0.63                & 0.17                  & \textbf{0.83}        & \textbf{1.09}      & \textbf{1.02}     \\
			CLIP4CIR~\cite{baldrati2022effective} & \textbf{0.47}             & \textbf{ 0.48}           & \textbf{0.45}               & \textbf{0.50}               & 0.18               & 0.65                & 0.48              & \textbf{0.51}     & \textbf{0.50}      & \textbf{0.65}    & \textbf{0.71}       & \textbf{0.27}         & 0.31                 & 0.69               & 0.82              \\

			\bottomrule
		\end{tabular}

		\begin{tabular}{l|c c c|c c c c|c c c c|c c c c}
			\multicolumn{1}{c}{}                  & \multicolumn{3}{c}{Noise} & \multicolumn{4}{c}{Blur} & \multicolumn{4}{c}{Weather} & \multicolumn{4}{c}{Digital}                                                                                                                                                                                                                                          \\

			\toprule
			\textbf{FashionIQ-C Toptee}           & \scriptsize{Gauss.}       & \scriptsize{Shot}        & \scriptsize{Impluse}        & \scriptsize{Defocus}        & \scriptsize{Glass} & \scriptsize{Motion} & \scriptsize{Zoom} & \scriptsize{Snow} & \scriptsize{Frost} & \scriptsize{Fog} & \scriptsize{Bright} & \scriptsize{Contrast} & \scriptsize{Elastic} & \scriptsize{Pixel} & \scriptsize{JPEG} \\



			\midrule
			TIRG~\cite{vo2019composing}           & 0.30                      & 0.28                     & 0.25                        & 0.36                        & 0.24               & 0.63                & 0.58              & 0.32              & 0.27               & 0.39             & 0.58                & 0.10                  & 0.69                 & 0.88               & 0.88              \\
			MAAF~\cite{dodds2020modality}         & 0.30                      & 0.28                     & 0.27                        & 0.45                        & 0.24               & 0.71                & 0.52              & 0.28              & 0.23               & 0.28             & 0.56                & 0.14                  & 0.52                 & 0.88               & 0.88              \\
			ARTEMIS~\cite{delmas2022artemis}      & 0.21                      & 0.23                     & 0.18                        & 0.37                        & 0.28               & 0.68                & 0.57              & 0.33              & 0.25               & 0.38             & 0.53                & 0.13                  & 0.66                 & 0.88               & 0.82              \\

			FashionViL~\cite{han2022fashionvil}   & 0.28                      & 0.28                     & 0.27                        & 0.44                        & \textbf{0.32}      & \textbf{0.85}       & \textbf{0.69}     & 0.38              & 0.33               & 0.36             & 0.69                & 0.15                  & \textbf{0.88}        & \textbf{1.09}      & \textbf{1.06}     \\
			CLIP4CIR~\cite{baldrati2022effective} & \textbf{0.42}             & \textbf{ 0.4}            & \textbf{0.42}               & \textbf{0.58}               & 0.21               & 0.76                & 0.49              & \textbf{0.46}     & \textbf{0.44}      & \textbf{0.60}    & \textbf{ 0.71}      & \textbf{0.24}         & 0.39                 & 0.78               & 0.84              \\

			\bottomrule
		\end{tabular}

		\label{tab:fashion_recall10_cat}
		\vspace{-0.3cm}}
\end{table}

\begin{table}[t!]
	\centering
	\small
	\caption{\sst{Recall@10 for text-image composed retrieval under 15 natural image corruptions in FashionIQ-C.}}
	{\setlength\tabcolsep{1.5pt}

		\sst{
			\begin{tabular}{l|c|c c c|c c c c|c c c c|c c c c}
				\multicolumn{2}{c}{}                  & \multicolumn{3}{c}{Noise} & \multicolumn{4}{c}{Blur} & \multicolumn{4}{c}{Weather} & \multicolumn{4}{c}{Digital}                                                                                                                                                                                                                                                                 \\

				\toprule
				\textbf{FashionIQ-C}                  & \scriptsize{Clean}        & \scriptsize{Gauss.}      & \scriptsize{Shot}           & \scriptsize{Impluse}        & \scriptsize{Defocus} & \scriptsize{Glass} & \scriptsize{Motion} & \scriptsize{Zoom} & \scriptsize{Snow} & \scriptsize{Frost} & \scriptsize{Fog} & \scriptsize{Bright} & \scriptsize{Contrast} & \scriptsize{Elastic} & \scriptsize{Pixel} & \scriptsize{JPEG} \\
				\midrule



				TIRG~\cite{vo2019composing}           & 23.8                      & 6.6                      & 6.1                         & 5.4                         & 8.1                  & 5.3                & 14.6                & 13.5              & 7.7               & 6.3                & 8.8              & 13.8                & 3.0                   & 15.2                 & 20.3               & 20.2              \\
				MAAF~\cite{dodds2020modality}         & 23.4                      & 7.2                      & 6.4                         & 5.9                         & 10.4                 & 5.0                & 15.8                & 12.3              & 6.6               & 5.5                & 7.3              & 12.7                & 3.1                   & 12.7                 & 19.4               & 19.4              \\
				ARTEMIS~\cite{delmas2022artemis}      & 24.9                      & 5.8                      & 6.0                         & 4.9                         & 9.4                  & 6.5                & 16.4                & 14.9              & 9.0               & 6.2                & 9.4              & 13.9                & 3.5                   & 16.2                 & 21.4               & 21.5              \\

				FashionViL~\cite{han2022fashionvil}   & 23.4                      & 6.1                      & 6.5                         & 5.9                         & 9.5                  & 7.2                & 19.3                & 15.8              & 7.8               & 7.3                & 8.0              & 16.5                & 3.5                   & 20.3                 & 25.7               & 24.9              \\
				CLIP4CIR~\cite{baldrati2022effective} & 35.9                      & 15.9                     & 15.2                        & 15.6                        & 19.4                 & 7.5                & 25.7                & 17.8              & 16.5              & 15.6               & 21.5             & 25.2                & 8.2                   & 13.3                 & 26.5               & 29.7              \\
				\bottomrule
			\end{tabular} }
		\label{tab:fashion_recall10_acc}
		\vspace{-0.3cm}}
\end{table}

	%

	\subsection{Analysis of COCO with image corruptions}

	To evaluate the compared models on more general domain, we implement our image  corruptions on the validation set of COCO~\cite{lin2014microsoft}, represented by CIRR-C.
	We set masked bounding box as the reference image, the raw image as the target image, and the labels of objects as modified text the following ~\cite{neculai2022probabilistic,saito2023pic2word}.
	The three compared models are trained on the CIRR dataset and evaluated on the validation set of COCO with 5000 images.
	The results show that large pretrained model CLIP4CIR have higher robustness than smaller models TIRG and ARTEMIS, which follow the same conclusion in paper Section 4.1.

\begin{table}[t!]
	\centering
	\small
	\caption{\sst{Relative robustness score for text-image composed retrieval under 15 natural image corruptions in COCO-C Recall@10.}}
	{\setlength\tabcolsep{2.2pt}

		\sst{
			\begin{tabular}{l|c c c|c c c c|c c c c|c c c c}
				\multicolumn{1}{c}{}                  & \multicolumn{3}{c}{Noise} & \multicolumn{4}{c}{Blur} & \multicolumn{4}{c}{Weather} & \multicolumn{4}{c}{Digital}                                                                                                                                                                                                                                          \\

				\toprule
				\textbf{COCO-C}                       & \scriptsize{Gauss.}       & \scriptsize{Shot}        & \scriptsize{Impluse}        & \scriptsize{Defocus}        & \scriptsize{Glass} & \scriptsize{Motion} & \scriptsize{Zoom} & \scriptsize{Snow} & \scriptsize{Frost} & \scriptsize{Fog} & \scriptsize{Bright} & \scriptsize{Contrast} & \scriptsize{Elastic} & \scriptsize{Pixel} & \scriptsize{JPEG} \\
				\midrule

				TIRG~\cite{vo2019composing}           & 0.19                      & 0.21                     & 0.14                        & 0.42                        & 0.25               & 0.62                & 0.58              & 0.35              & 0.21               & 0.51             & 0.89                & 0.05                  & 0.40                 & 0.40               & 0.72              \\
				ARTEMIS~\cite{delmas2022artemis}      & 0.14                      & 0.16                     & 0.08                        & 0.43                        & 0.22               & 0.72                & 0.52              & 0.41              & 0.32               & 0.45             & 1.06                & 0.05                  & 0.40                 & 0.48               & 0.70              \\
				CLIP4CIR~\cite{baldrati2022effective} & 0.52                      & 0.58                     & 0.52                        & 0.65                        & 0.12               & 0.85                & 0.36              & 0.51              & 0.49               & 0.71             & 0.90                & 0.10                  & 0.24                 & 0.77               & 0.77              \\
				\bottomrule
			\end{tabular} }
		\label{tab:coco_recall10_acc_img}
		\vspace{-0.3cm}}
\end{table}

\section{More Related Works}
\label{supp:related_work}
\textbf{Diagnostic analysis.}
Recently, a range of benchmarks for visual understanding have been proposed, including datasets for image captioning~\cite{shekhar2017foil}, visual question answering~\cite{johnson2017clevr}, visual reasoning~\cite{zerroug2022benchmark} and visio-linguistic compositional reasoning\sst{~\cite{thrush2022winoground,yuksekgonul2022and,ma2023crepe}}.
\sst{
	For text-image composed retrieval, the benchmarks can be categorized into sythetic-based datasets by cubes~\cite{vo2019composing} or natural scenes~\cite{gu2023compodiff}, fashion-based datasets~\cite{han2017automatic,berg2010automatic,guo2019fashion}, object-state dataset~\cite{isola2015discovering} and open domain dataset~\cite{liu2021image}.
	Among them, the majority of the textual descriptions are limited by predefined attributes~\cite{han2017automatic,vo2019composing,isola2015discovering}.
	To overcome this limitation, FashionIQ~\cite{guo2019fashion} and CIRR~\cite{liu2021image} leverage the flexibility of natural language and becomes the most widely used benchmarks in fashion domain and open domain respectively.}
We expand and categorize the test set of CIRR benchmark from its main and unused extended annotation to probe specific text understanding in numerical variation, attribute variation, object removal, background variation, and fine-grained variation.
%
Similar to ours, many diagnostic datasets are also synthetic ones.
CLVER~\cite{johnson2017clevr} is a synthetic dataset to probe elementary vision reasoning including color, shape, and spatial relationships.
%
%
CVR~\cite{zerroug2022benchmark} generates irregular shape, location, color, etc, and designed for detecting the outlier from a small set of generated images.
However, these are all simulated images and are not generated by imitating natural scenes.
CasualVQA~\cite{agarwal2020towards} and CompoDiff~\cite{gu2023compodiff} generate images imitating natural scenes.
However, CasualVQA is designed for visual question answering tasks and the generated images include noticeable artifacts.
While ComoDiff is designed for text-image composed retrieval, but they generate by replacing the objects (noun) only instead of attributes (numeral, adjective, manipulation instruction) like ours, so cannot pinpoint the target reasoning abilities.
\sst{
	To detect the compositional abilities, visio-linguistic compositional reasoning diagnostic benchmarks supply variation of objects~\cite{ma2023crepe,yuksekgonul2022and}, attributes~\cite{ma2023crepe,yuksekgonul2022and}, or text order~\cite{yuksekgonul2022and,thrush2022winoground}.
	However, these benchmarks only supply image-text pairs for single-modality queries instead of image-text-image triplets for multi-modality queries.
	Also their text composition~\cite{ma2023crepe,yuksekgonul2022and} may not have corresponding image like `grass eat horse'.
	In comparision, our CIRR-D supplies the first visio-linguistic composition reasoning benchmark for text-image composed retrivel task in natural scenes.}

\textbf{Text-image composed retrieval.}
Composed image retrieval aims to retrieve the target image, where the input query is specified in the form of an image plus other interactions, such as relative attribute\cite{parikh2011relative}, natural language\cite{chen2020image, vo2019composing}, spatial layout\cite{mai2017spatial}, to describe the desired modifications.
Among them, natural language as the most pervasive interaction between humans and computers to convey intricate specification has attracted increasing attention, which has often led `\textit{composed image retrieval}' to become interchangeable with `\textit{text-guided image retrieval}' in the literature.
We term the task as text-image composed retrieval to clarify the composition of the query.
Traditional text-image composed retrieval models implement separate independent image and text encoders, whose features are combined with late fusion.
For example, TIRG~\cite{vo2019composing} and Artemis~\cite{delmas2022artemis} implement separate pre-trained ResNet as image encoder and LSTM as text encoder.
Until recently, with the power of unified multimodal space CLIP~\cite{radford2021learning}, current text-image composed retrieval models achieved a noticeable improvement.
For example, CLIP4CIR~\cite{baldrati2022effective} implements a light adapter as image-text late fusion and further tune it in target domains.
Further based on CLIP, FAME~\cite{han2023fame} and CASE~\cite{levy2023data} separately  implement early cross attention between text and image, which shows obvious improvement.

\section{Limitation}
\label{sec:limitation}
\begin{figure}[t]
	\centering
	\includegraphics[width=1.0\textwidth]{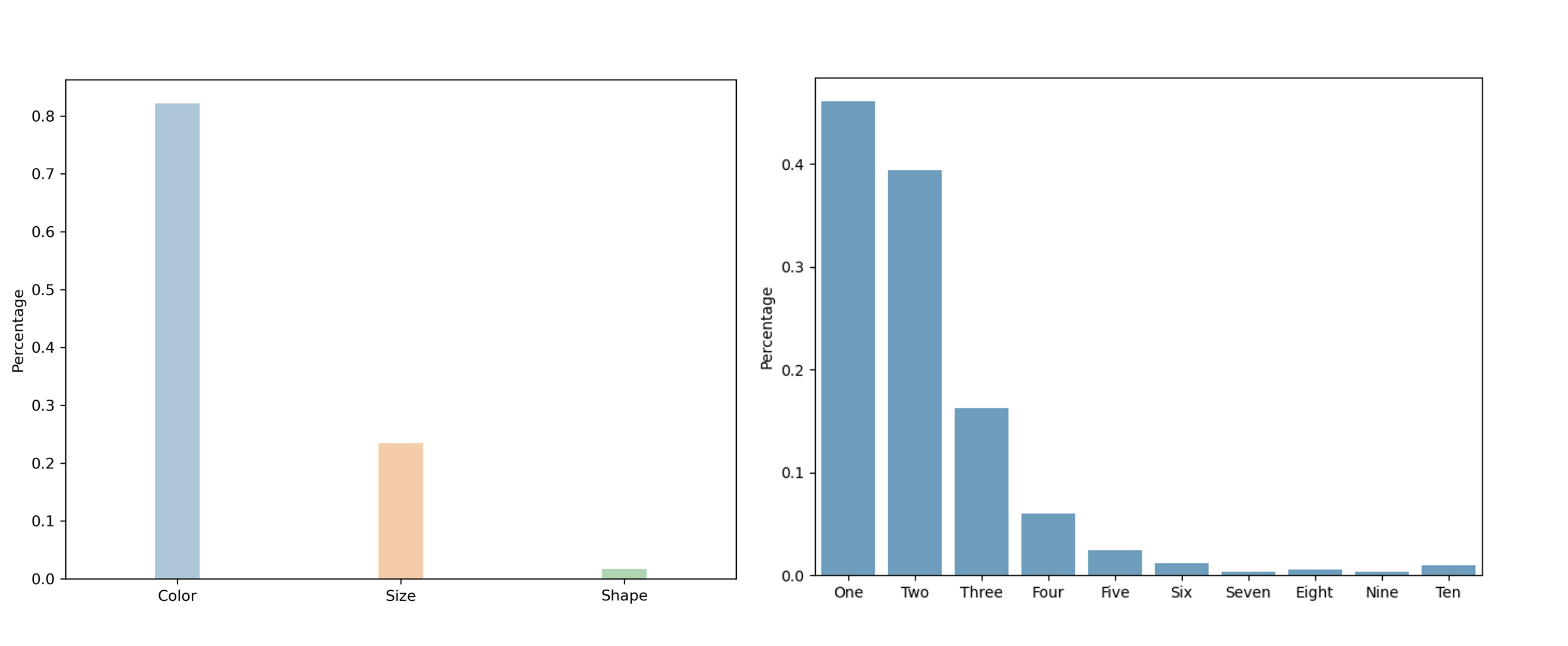}
	\caption{CIRR-D distribution for attribute variants and numerical variations.}
	\label{fig:distribution}

\end{figure}
\begin{figure}[t]
	\centering
	\includegraphics[width=1.0\textwidth]{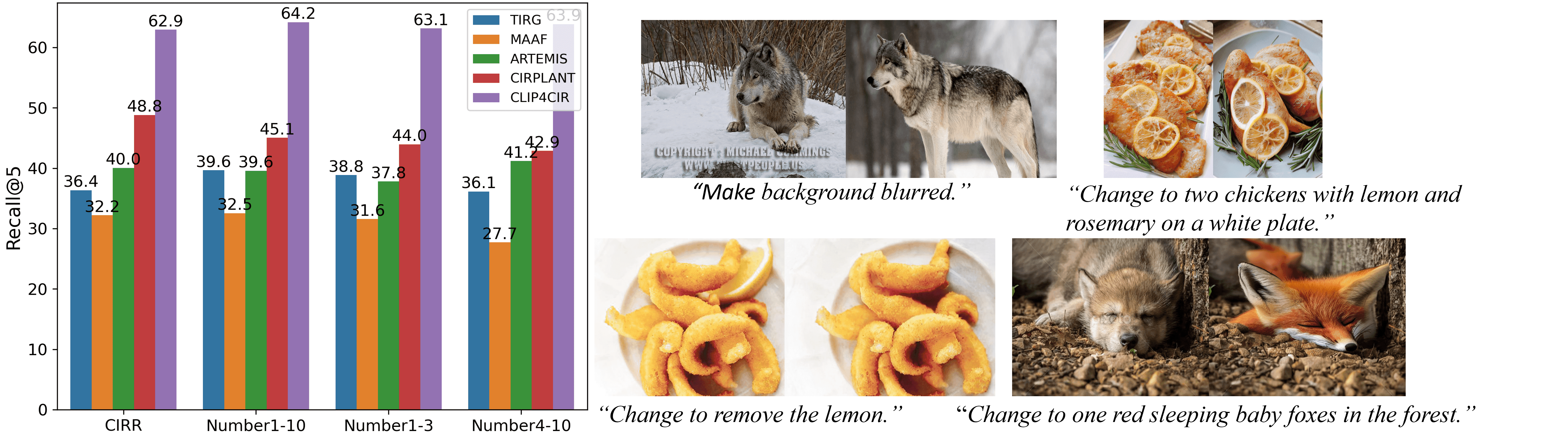}
	\caption{Left: Recall@5 on CIRR-D numerical queries. Right: Samples visualization of CIRR-D. The left and right images are the reference and target images. Except the upper left triplet, rest target images are synthetic.} \vspace{-0.3cm}
	\label{fig:cirr-D}
\end{figure}

We discuss the limitation of the proposed benchmarks in this section.
For benchmarking natural corruption in CIRR-C and FashionIQ-C, the method of simulating real-world corruption with the noise still has limitations.
For benchmarking textual understanding in CIRR-D, it has long-tail distribution.
As shown in Figure.~\ref{fig:distribution}, both numerical and attribute evaluation set follows the long-tail distribution.
The numerical set has a large number of samples in the range of 1 to 3, while each category from 4 to 10 has only a small number of samples.
The attribute evaluation set has a large number of samples with colour variations and a small number of samples with size variations.
The imbalanced distribution can lead to bias towards the categories with more data.
Further, we visualise the performance of the query with number one to three, four to ten respectively shown in Figure~\ref{fig:cirr-D} left.
\sst{The average recall@5 of five evaluated methods are 43.06\% on number one to three, 42.36\% on numbers four to ten and 44.2\% on number one to ten respectively. (A sentence with multiple numbers will be categorized to multiple categories, thus number one to three and number four to ten can overlap.) Based on this subtle accuracy change, we speculate that the model also possesses a similar capability for recognizing the less frequent samples (number four to ten) in the long-tail distribution as it does for the more frequent samples (number one to three).}

\section{License}
All the models in this study are available to the public.
The model code for TIRG~\cite{vo2019composing} and MAAF~\cite{dodds2020modality} have the Apache License Version 2.0, ARTEMIS~\cite{delmas2022artemis} has CC BY-NC-SA 4.0 License, CIRPLANT~\cite{liu2021image} has MIT license and FashionViL~\cite{han2022fashionvil} has BSD License.
We will provide CIRR-C, FashionIQ-C and CIRR-D publicly.
These datasets are based on existing CIRR~\cite{liu2021image} and FashionIQ~\cite{guo2019fashion}.
For CIRR-C and FashionIQ-C, we didn't add any new images or text sources.
For CIRR-D, we further generate synthetic images and text to expand the original CIRR dataset.
All of these datasets are available to the public and we apply similar licenses to our testbed code and our proposed benchmarks.

\end{document}